\newcommand{\CLASSINPUTbaselinestretch}{0.97} 
\newcommand{\CLASSINPUTinnersidemargin}{0.62in} 
\newcommand{\CLASSINPUToutersidemargin}{0.62in} 

\newcommand{\trtitle}{Affect-Driven Modelling of Robot Personality for Collaborative Human-Robot Interactions}

\documentclass[journal,twoside,nocompress,letterpaper]{IEEEtran}
\ifCLASSOPTIONcompsoc
  \usepackage[nocompress]{cite}
\else
  \usepackage{cite}
\fi
\usepackage{graphicx}
\usepackage{todonotes}

\usepackage{amsfonts}
\usepackage{amsmath}
\usepackage{amssymb}
\usepackage{amsthm}
\usepackage{enumitem}
\usepackage{algorithmic}
\usepackage{xcolor}
\usepackage{array}

\ifCLASSOPTIONcompsoc
  \usepackage[caption=false,font=footnotesize,labelfont=sf,textfont=sf]{subfig}
\else
  \usepackage[caption=false,font=footnotesize]{subfig}
\fi
\usepackage{adjustbox}
\usepackage{tcolorbox}
\usepackage{diagbox}

\ifCLASSOPTIONcaptionsoff
  \usepackage[nomarkers]{endfloat}
 \let\MYoriglatexcaption\caption
 \renewcommand{\caption}[2][\relax]{\MYoriglatexcaption[#2]{#2}}
\fi
\usepackage{caption}
\usepackage{tabularx}
\usepackage{booktabs}                   
\usepackage{multicol}                   
\usepackage{multirow}                   
\usepackage{makecell}
\usepackage{rotating}                   

\usepackage{url}
\usepackage{bm}

\usepackage{anyfontsize}
\usepackage{t1enc}

% ...

\ifCLASSINFOpdf
  \usepackage[pdftex]{thumbpdf}
\else
  \usepackage[dvips]{thumbpdf}
\fi

\newcommand\MYhyperrefoptions{bookmarks=true,bookmarksnumbered=true,
pdfpagemode={UseOutlines},plainpages=false,pdfpagelabels=true,
colorlinks=true,linkcolor={blue},citecolor={green},urlcolor={black},
pdftitle={RobotPersonalityforCollaborativeHRI},
pdfsubject={SocialRobotics},
pdfauthor={Nikhil Churamani},
pdfkeywords={Affective Computing, Multi-modal Perception, Self-organisation, Robot Personality, Social Robotics, Neural Networks, Deep Learning}}
\ifCLASSINFOpdf
\usepackage[\MYhyperrefoptions,pdftex]{hyperref}
\else
\usepackage[\MYhyperrefoptions,breaklinks=true,dvips]{hyperref}
\usepackage{breakurl}
\fi
\usepackage[nolist]{acronym}
\usepackage{balance}
\begin{document}

\title{\trtitle}
\author{Nikhil~Churamani\IEEEauthorrefmark{1}\href{https://orcid.org/0000-0001-5926-0091}{$^{\includegraphics[scale=0.5]{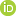}}$}, 
      Pablo~Barros\IEEEauthorrefmark{2}\href{https://orcid.org/0000-0002-6517-682X}{$^{\includegraphics[scale=0.5]{Figures/orcid.png}}$}, 
        Hatice~Gunes\IEEEauthorrefmark{1}\href{https://orcid.org/0000-0003-2407-3012}{$^{\includegraphics[scale=0.5]{Figures/orcid.png}}$}, 
      and~Stefan~Wermter\IEEEauthorrefmark{3}\href{https://orcid.org/0000-0003-1343-4775}{$^{\includegraphics[scale=0.5]{Figures/orcid.png}}$} 
\IEEEcompsocitemizethanks{
\IEEEcompsocthanksitem \IEEEauthorrefmark{1}N. Churamani and H. Gunes are with the Department of Computer Science and Technology, University of Cambridge, UK. \protect\\ {\tt Email: \{nikhil.churamani, hatice.gunes\}@cl.cam.ac.uk}
\IEEEcompsocthanksitem \IEEEauthorrefmark{2}P. Barros is with the Cognitive Architecture for Collaborative Technologies (CONTACT) Unit, Istituto Italiano di Tecnologia, Genova, Italy. \protect\\
{\tt Email: pablo.alvesdebarros@iit.it}
\IEEEcompsocthanksitem \IEEEauthorrefmark{3}S. Wermter is with Knowledge Technology (WTM), Department of Informatics, University of Hamburg, Germany. \protect \\ 
{\tt Email: wermter@informatik.uni-hamburg.de}
\IEEEcompsocthanksitem {\footnotesize This work is supported by the German Research Foundation (DFG) under project CML (TRR 169), the European Union under projects SECURE~(No.~642667) and SOCRATES~(No.~721619), and the NVIDIA Corporation. N. Churamani and H. Gunes are supported by EPSRC under grants: EP/R513180/1 (ref.2107412) and EP/R030782/1. The authors also thank Sascha Griffiths, Mathias Kerzel, Sven Magg, and German Parisi for their insights.}

}

}


\IEEEtitleabstractindextext{
\begin{abstract}

 Collaborative interactions require social robots to adapt to the dynamics of human affective behaviour. Yet, current approaches for affective behaviour generation in robots focus on instantaneous perception to generate a one-to-one mapping between observed human expressions and static robot actions. In this paper, we propose a novel framework for personality-driven behaviour generation in social robots. The framework consists of (i)~a hybrid neural model for evaluating facial expressions and speech, forming intrinsic affective representations in the robot, (ii)~an \textit{Affective Core}, that employs self-organising neural models to embed robot personality traits like \textit{patience} and \textit{emotional actuation}, and (iii)~a Reinforcement Learning model that uses the robot’s affective appraisal to learn interaction behaviour. For evaluation, we conduct a user study (n $=31$) where the \acs{NICO} robot acts as a \textit{proposer} in the Ultimatum Game. The effect of robot personality on its negotiation strategy is witnessed by participants, who rank a \textit{patient} robot with \textit{high emotional actuation} higher on \textit{persistence}, while an \textit{inert} and \textit{impatient} robot higher on its \textit{generosity} and \textit{altruistic} behaviour.

\end{abstract}

}

\maketitle

\IEEEdisplaynontitleabstractindextext

%
\IEEEpeerreviewmaketitle

\ifCLASSOPTIONcompsoc
\IEEEraisesectionheading{
\section{Introduction}
\label{sec:introduction}}
\else
\section{Introduction}
\label{sec:introduction}
\fi




\IEEEPARstart{I}{n} collaborative \ac{HRI} scenarios, where robots need to effectively negotiate with humans, it is particularly important for them to be sensitive to human affective behaviour~\cite{breazeal2003emotion}. Furthermore, instead of using static behaviour policies that fail to engage users over continued interactions~\cite{Leite2013Social}, robots should understand the affective impact of their interactions, and, over time, evolve their behaviour. 

Much of the current research in Affective Computing and Social Robotics focuses on instantaneous (frame-based or using very-short sequences) affect perception (see~\cite{ Sariyanidi2015Automatic, Corneanu2016Survey} for an overview). Although this works well for short-term interactions, longer context-driven conversations require a robot to analyse and understand human behaviour over an entire interaction~\cite{KIRBY2010Affective}. This is primarily because frame-based techniques rely on glimpses of heightened audiovisual stimuli to infer the affective state of the user~\cite{Sariyanidi2015Automatic}, missing out subtle nuances in expressions. To engage the users, robots should form an evolving understanding of human behaviour~\cite{baxter2011long} by modelling affective representations~\cite{scherer2000psychological,KIRBY2010Affective} that track user behaviour over time and create a dynamic and robust model for its affective appraisal. Additionally, for naturalistic interactions, developing idiosyncratic behavioural tendencies can provide means for embedding specific personality traits in robots to influence their affective appraisal as well as their behaviour~\cite{Han2013Robotic}. 

Humans are particularly known to possess such innate behavioural traits that affect their experiences. The \textit{affective core}~\cite{EMDE1991251,Russel2003Core} of an individual acts as a primitive pleasure model modulating \textit{agency} and \textit{intrinsic motivation}. It impacts an individual’s behaviour, given their temperament, that is, an inherent inclination that shapes up an individual's behaviour~\cite{Rothbart2000Temperament}, influencing their subjective appraisal of the surroundings~\cite{Thomas1970Origin} as well as their decision-making~\cite{BANDYOPADHYAY2013Role}. Modelling such an \textit{affective core} for robots can provide the necessary modulation for personality-driven behaviour learning.

Collaborative \ac{HRI} scenarios also require modelling naturalistic interaction dynamics between humans and robots. Hence, achieving \textit{adaptability}, such that a robot shows an improved and evolving understanding of user behaviour, becomes a principal objective. Exhibiting such `personal ontogeny'~\cite{Robins2005Sustaining} can also hint at the robot intelligently interacting with users. This work proposes the robot's affective appraisal as the basis of learning adaptive interaction behaviours. Different from existing approaches that mimic the user's expressions~\cite{Churamani2017Teaching,Paiva2017EVA}, here we propose forming evolving intrinsic affective responses in the robot towards a user's affective state, providing for personalised and adaptive interaction capabilities. These affective responses are modelled as the robot's \textit{affective memory}~\cite{Barros2017ASelf}, that is, the affective impact of past interactions with a user, as well as its \textit{mood}~\cite{Churamani2018Learning,2018arXiv180800252B}, representing its affective appraisal. This affective appraisal is modulated by specifically modelled intrinsic personality traits, or the \textit{affective core} of the robot to learn appropriate robot behaviour while negotiating resources with users in the \textit{Ultimatum Game}~\cite{Guth1982An}. The main contributions of this work can be summarised as the following: 

\begin{enumerate}[leftmargin=0.4cm]

    \item A deep, hybrid neural model is trained for robust multi-modal affective appraisal, evaluating the user's facial expressions and speech. These evaluations help form the \textit{affective memory} and the intrinsic \textit{mood} of the robot. 

    \item An \textit{Affective Core} for the robot, modelled using recurrent self-organising neural networks, is proposed to enforce distinct personality dispositions on the \textit{mood} of the robot. Two influences, namely, `time perception' as the impact of the duration of interaction, and the robot's `social conditioning' or emotional actuation, that is, the intensity with it experiences an interaction, are explored.

    \item Robot's \textit{mood} is then used to learn an optimal negotiating behaviour in the Ultimatum Game~\cite{Guth1982An}. An actor-critic-based~\cite{Lillicrap2015Continuous} \ac{RL} model is proposed that learns to negotiate resources with users based on their affective responses to the robot's offers.

\end{enumerate}

The multi-modal appraisal allows the robot to comprehend evolving human affective behaviour. The emergence of different personality dispositions, as a result of the \textit{affective core} of the robot, modulate its intrinsic \textit{mood}, forming the basis for learning robot behaviour. The Ultimatum Game is explored as it underlines the expectations from robots in collaborative \ac{HRI} scenarios, particularly concerning adaptability and naturalistic interactions.

\section{Related Work}
\label{sec:related}

The affective impact of one's interactions with others plays an important role in human cognition~\cite{jeon2017emotions}. The \textit{core affect} in an individual, forms a neurophysiological state~\cite{Russel2003Core} resulting from the interplay between the valence of an experience and the emotional arousal it invokes. This influences how people perceive situations and regulates their responses. From early stages of development, human behaviour is seen to be governed by such an \textit{affective core}~\cite{EMDE1991251} that develops, initially, as a procedural understanding of their surroundings, and later, to a more cognitive representation that influences human \textit{agency} and \textit{behaviour}. Self-regulatory aspects of \textit{personality} acquired as a result of interactions are essential for cognitive development and act as anchors for perception and understanding. Such individualistic attributes of \textit{temperament}, evolving into \textit{personality}, can be seen as the ``basis for dispositions and orientations towards others and the physical world and for shaping the person's adaptations to that world''~\cite{Rothbart2000Temperament}.

Understanding the evolution of human affective behaviour enables us to emulate such characteristics in social robots. It allows robots to ground intrinsic models of affect to improve their interaction capabilities. This section presents a brief overview on multi-modal affective appraisal (Section~\ref{subsec:affagent}) and behaviour synthesis (Section~\ref{subsec:affbehave}) in social robots discussing different existing frameworks that use affective appraisal as the basis for modelling robot behaviour in \ac{HRI} scenarios.

\vspace*{-1mm}
\subsection{Multi-modal Affective Appraisal in Social Agents}
\label{subsec:affagent}

Humans interact with each other using different verbal and non-verbal cues such as facial expressions, gestures and speech. Although various \textit{outward} signals~\cite{Gunes2011Emotion} can be observed to model emotion perception in agents, here we discuss facial expressions and auditory signals as the predominantly used modes of perception, and how these can be combined for multi-modal affect perception in agents. 

\subsubsection{\acf{FER}}
Evaluating facial expressions is one of the most straightforward and commonly used approach for affect perception. Facial expressions can be categorised into several emotional categories~\cite{Ekman1971Constants} or represented on a dimensional scale~\cite{Gunes2011Emotion,Yannakakis2017The}. Traditionally, computational models have used hand-crafted features such as shape-based, spectral or histogram-based analysis, and other feature-based transformations for affect perception (see~\cite{zeng2009survey,Sariyanidi2015Automatic, Corneanu2016Survey} for a detailed analysis). More recently, deep learning has enhanced the performance of \ac{FER} models by reducing the dependency on the choice of features and instead, learning these features directly from the data~\cite{Li2018Deep, Kollias2018IJCNN}. Although these work well in clean and noise-free environments, spontaneous emotion recognition in less controlled settings is still a challenge~\cite{Sariyanidi2015Automatic}. Thus, the focus has now shifted towards developing techniques that are able to recognise facial expressions in real-world conditions~\cite{Zafeiriou2017AFF,KOSSAIFI201723}, robust to movements of the observed person, noisy environments and occlusions~\cite{Zen2016Learning}.

\subsubsection{\acf{SER}}
Affective responses can also be evaluated using speech, either by processing spoken words to extract the sentiment behind them or understanding speech intonations. While spoken words convey meaning, \textit{paralinguistic} cues enhance a conversation by highlighting the affective motivations behind these spoken words~\cite{Gunes2010ADC}. Despite providing information about the context and intent~\cite{WHISSELL1989113} in an interaction, it is difficult to deduce the emotional state of the individual using only linguistic information~\cite{Furnas1987}. Extracting spectral and prosodic representations can help better analyse affective undertones in speech. Different studies on \ac{SER} (see~\cite{Schuller2018SER} for an overview) make use of representations such as \acf{MFCC} or features like \textit{pitch} and \textit{energy} to evaluate expressed emotions. More recently, (deep) learning is employed to extract relevant features directly from the raw audio signals~\cite{Keren2016Convolutional,Tzirakis2017End}.

\subsubsection{Combining Modalities}
As certain emotions are better expressed using facial expressions (or body gestures) while others are elucidated in speech~\cite{sebe2005multimodal}, considering behavioural cues across multiple modalities has shown to improve the perception capabilities of agents~\cite{schels2013multi}. Most of the current approaches~\cite{Tzirakis2017End,PORIA201798} combine different modalities to recognise the emotions expressed by an individual. This combination can either be achieved using weighted averaging or majority voting~\cite{schels2013multi} from individual modalities~\cite{busso2004analysis} or feature-based sensor fusion~\cite{Tzirakis2017End,Kahou2016EmoNets} and deep learning~\cite{PORIA201798}. 

\subsubsection{Intrinsic Representation of Affect}
For long-term adaptation, it is important that robots not only recognise affect but also model \textit{continually} evolving intrinsic affective representations~\cite{paiva2014emotion}. Kirby~et~al.~\cite{KIRBY2010Affective} explore slow-evolving affect models such as \textit{moods} and \textit{attitudes} that consider personal history and the environment to estimate an affective state for the robot. Barros~et~al.~\cite{2018arXiv180800252B} propose the formation of an intrinsic \textit{mood} that uses an \textit{affective memory}~\cite{Barros2017ASelf} of an individual as an influence over the spontaneous perception. The WASABI model~\cite{Becker-Asano2009WASABI} represents the intrinsic state of the robot on a \textit{\acs{PAD}}-scale that adapts as the agent interacts with the user. In the SAIBA framework~\cite{Le2011Design}, the agent's intrinsic state is modelled using mark-up languages that models intent in the robot and uses it to generate corresponding agent behaviour. The \acs{(DE)SIRE} framework~\cite{lim2012towards} represents this intrinsic affect as a vector in a $4$-d space for the robot which is then mapped to corresponding expressions across different modalities. Schr\"oder~et~al.~\cite{Schroder2012Building} create four different virtual characters to measure user engagement, with their behaviour corresponding to a particular quadrant of the arousal-valence space. Each character thus tries to invoke the same responses in the users as its inclination. Although all these approaches are able to provide necessary biases for modelling intrinsic personalities in agents, they require careful initialisation, across $n$-dimensional vector-spaces, to result in the desired effect. It will be beneficial if these intrinsic representations could be learnt dynamically by the agent as a result of its interactions.

\begin{figure*}[t]
\centering
\includegraphics[width=1.0\textwidth]{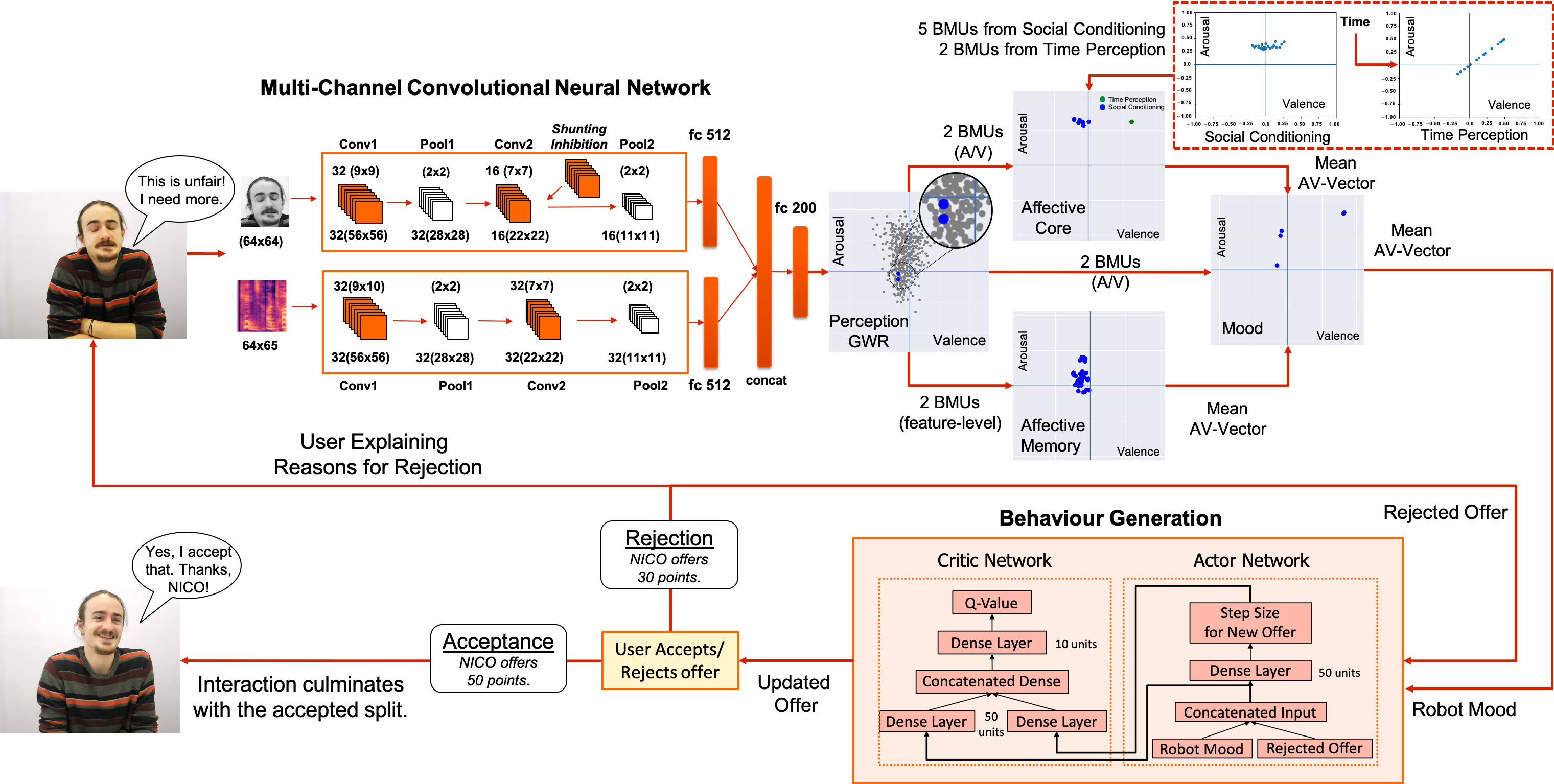}
\caption{Proposed Framework: The \textbf{\acs{MCCNN}} network combines facial and auditory features. The \textbf{Perception-\acs{GWR}} creates prototypes of this combined representation with the \textit{winner} neurons encoding arousal-valence. Repeated interactions form the \textbf{Affective Memory}. The \textbf{Affective Core} models personality traits in the robot, resulting from its \textit{social conditioning} and \textit{time perception}. Current perception, affective memory and the affective core influence \textbf{Mood} formation. Robot's mood is used by the \textbf{Behaviour Generation} model to learn negotiating behaviour in the \textit{Ultimatum Game}.}
\label{fig:mmemotion}
\vspace{-2mm}
\end{figure*}

\subsection{Behaviour Synthesis in Social Agents}
\label{subsec:affbehave}

Recent works on behaviour learning in social agents investigate the role of affect as a motivation to interact with their environment. Such strategies may include affective modulation on computation of the reward function where explicit feedback from the user is shown to speed up learning~\cite{Broekens2007Emotion}. Alternatively, affective appraisal can be viewed as an inherent quality of the robot, motivating it to interact with its environment~\cite{Han2013Robotic,Moerland2018}. Affect is modelled as an evaluation of physiological changes (changing battery level or motor temperatures) that occur in the robot, with their behaviour influenced by \textit{homeostatic} drives that lead towards a stable internal state~\cite{Konidaris2006}. Other approaches examine different cues such as \textit{novelty} and the \textit{relevance} of an action to the task to appraise the robot's performance~\cite{Sequeira2011}. In case of \textit{value-based} approaches~\cite{Jacobs2014}, the state-space of the robot is mapped onto different affective states and the value of any state represents the affective experience of the robot in that state. Reward-based approaches, on the other hand, consider \textit{temporal changes} in the reward or the reward itself as the basis of the robot experiencing different affective states~\cite{Ahn2005}.


Proposing a framework for modelling naturalistic interactions, this work attempts to move beyond expression recognition tasks towards grounding evolving affective representations in the robot that help estimate its role in an interaction. Adapting the robot's intrinsic state in response to changing human behaviour will allow for smoother transitions during interactions. This will help avoid the pitfalls of frame-based expression recognition techniques that facilitate only static robot behaviours. Furthermore, affective appraisal contributing, not just to the intrinsic state of the robot but also acting as a reward for appropriate behaviour, is expected to improve the robot's ability to comprehend user responses, providing an evaluation of its own performance.

%

\section{The Proposed Framework}
\label{sec:framework}

In this paper, we propose an affective framework for modelling robot personality and behaviour generation, consisting of four main components (see Fig.~\ref{fig:mmemotion}). Firstly, we present the multi-modal perception (Section~\ref{sec:perception}) that observes the facial expressions and speech intonations of the user to form intrinsic models of affect in the robot. Continuously tracking the user's affective state to model an \textit{affective memory}~\cite{Barros2017ASelf} of the user elicits an intrinsic response towards the user in the form of the robot's \textit{mood}. Secondly, the \textit{affective core} is proposed for the robot (see Section~\ref{sec:core}) that models personality dispositions on its affective appraisal. Section~\ref{sec:mood}, describes how the different intrinsic and extrinsic evaluations such as users' expressions, the \textit{affective memory} and the \textit{affective core} of the robot impact its \textit{mood}. This intrinsic \textit{mood} represents the robot's \textit{state} and forms the basis for the behaviour learning model. Section~\ref{sec:behavemodel} describes the \ac{RL} model that enables the robot to learn to effectively negotiate with users in the Ultimatum Game scenario.



\vspace*{-1mm}
\subsection{Multi-Modal Affect Perception}
\label{sec:perception}
The affect perception model, adapted from~\cite{Churamani2018Learning, 2018arXiv180800252B}, consists of three components, namely the \acf{MCCNN} network~\cite{barros2016developing} for multi-modal feature extraction and fusion, the Perception-\ac{GWR} for prototyping extracted features to improve the robustness of the model to changing lighting conditions and variance within an individual's expressions, and the \textit{affective memory}~\cite{Barros2017ASelf} that evaluates how the affective state of the user evolves during an interaction (see Fig~\ref{fig:mmemotion}) . 

\subsubsection{The \acs{MCCNN} Network}
\ac{MCCNN}~\cite{barros2016developing, Churamani2018Learning} consists of two separate channels for processing facial and auditory information and then combines the learnt features into a combined representation. Rather than using categorical labels for classification, the model is adapted to represent affect in the form of the \textit{valence} and \textit{arousal} dimensions (see Fig.~\ref{fig:mmemotion}).

The face channel takes a $(64\times64)$ greyscaled \textit{mean-face} image from every $12$ frames (considering a $500$ milliseconds window) recorded at $25$ FPS. It consists of $2$ convolutional (\textit{conv}) layers, each followed by $(2\times2)$ max-pooling. The first layer performs $(9\times9)$ convolutions while the second consists of $(7\times7)$ filters using \textit{shunting inhibition}~\cite{FREGNAC2003Shunting} to obtain filters robust to geometric distortions. The \textit{conv} layers are followed by a fully-connected (FC) layer consisting of $512$ units.

The audio channel uses Mel-spectrograms computed for every $500$ milliseconds of the audio signal, re-sampled to $16$ kHz and \textit{pre-emphasised}. A frequency resolution of $1024$ Hz is used, with a \textit{Hamming window} of $10$ms, generating Mel-spectrograms consisting of $64$ bins with $65$ descriptors each. The audio channel consists of two \textit{conv} layers with a filter size of $(9\times10)$ and $(7\times7)$ each followed by $(2\times2)$ max-pooling. The conv layers are followed by a FC layer with $512$ units.

The FC layers from both the face and audio channels are concatenated into a single dense representation consisting of $1024$ units and connected to another FC layer consisting of $200$ units. This enables the network to be trained to extract features that are able to predict arousal and valence values by combining the two modalities (see Section~\ref{sec:train_MCCNN}). 

\subsubsection{Perception-\acs{GWR}}
Even though the \ac{MCCNN} model can predict (using a 2-unit linear activation-based \ac{MCCNN} output layer) the affective state of the user in terms of the arousal and valence it encodes, variance within an individual's expressions may result in different outputs for the same affective state. Thus, to allow for a more robust approach, it is beneficial to adopt a developmental view on affect perception that can account for the variance with which users express the same affective state~\cite{barros2016developing}. We achieve this by using a \acf{GWR} network~\cite{MARSLAND20021041} that incrementally builds feature representations as the model receives different inputs, accounting for the variance in audio-visual stimuli (see~\cite{2018arXiv180800252B} for a detailed analysis). The extracted (\textit{fused}) feature representations from the $200$ unit FC layer are passed to the Perception-\ac{GWR} which learns feature prototypes, in an unsupervised manner, that represent the users' expressions over the two modalities (see Section~\ref{sec:train_GWR}). Thus, rather than considering the output of the \ac{MCCNN} classifier, we extract the learnt feature prototypes from the Perception-\ac{GWR} by taking the two \textit{winner} neurons closest to the input. These winner neurons are then classified using the \ac{MCCNN}-output layer into the encoded arousal-valence values. 

\subsubsection{Affective Memory}
To model long-term interactions, the robot needs to account for past interactions with users, forming a memory model that grows and adapts with them. Such an \textit{affective memory}~\cite{Barros2017ASelf} (see Fig.~\ref{fig:mmemotion}), developing as the robot interacts with the user, forms an expectation model for the robot that can reduce the impact of sudden changes in perception due to misclassifications or noise. As users interact with the robot, $2$ \acfp{BMU} or \textit{winner neurons} from the Perception-\ac{GWR} model, that is, feature prototypes for each $500$ milliseconds of audio-visual input, are used to train the robot's \textit{affective memory} (see Section~\ref{sec:train_GWR}). This memory is modelled using a Gamma-\ac{GWR} network~\cite{PARISI2017Lifelong} (explained in detail in Section~\ref{sec:core}) consisting of neurons with recurrent connections remembering past interactions. 
\begin{figure*}
\centering
\hspace{-0.5em}\subfloat[Patient ($\tau=0.01$) Decay.\label{fig:patient_time}]{\includegraphics[width=0.25\textwidth]{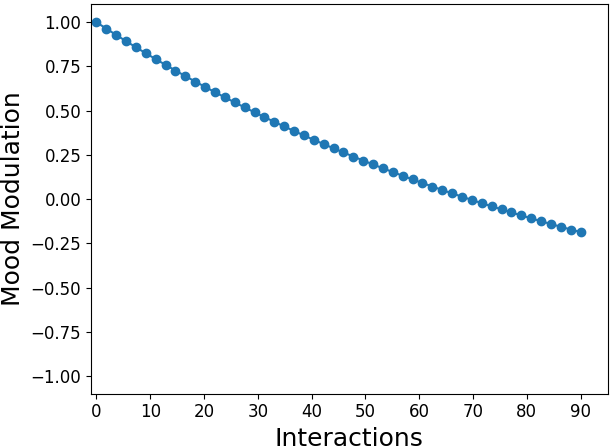}} \hspace{0.03em}
\subfloat[Impatient ($\tau=0.08$) Decay.\label{fig:impatient_time}]{\includegraphics[width=0.25\textwidth]{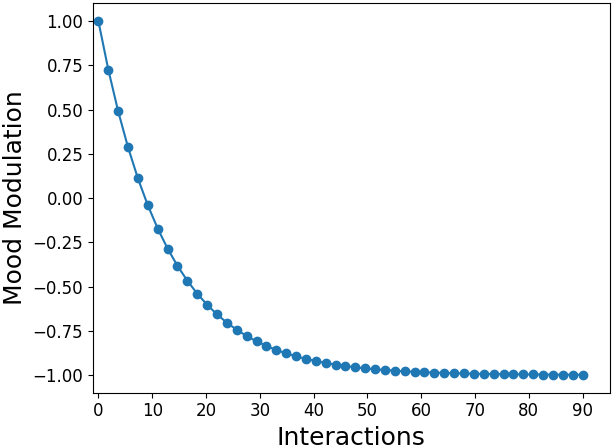}} \hspace{0.03em}
\subfloat[Patient Affective Core.\label{fig:patient_affcore}]{\includegraphics[width=0.25\textwidth]{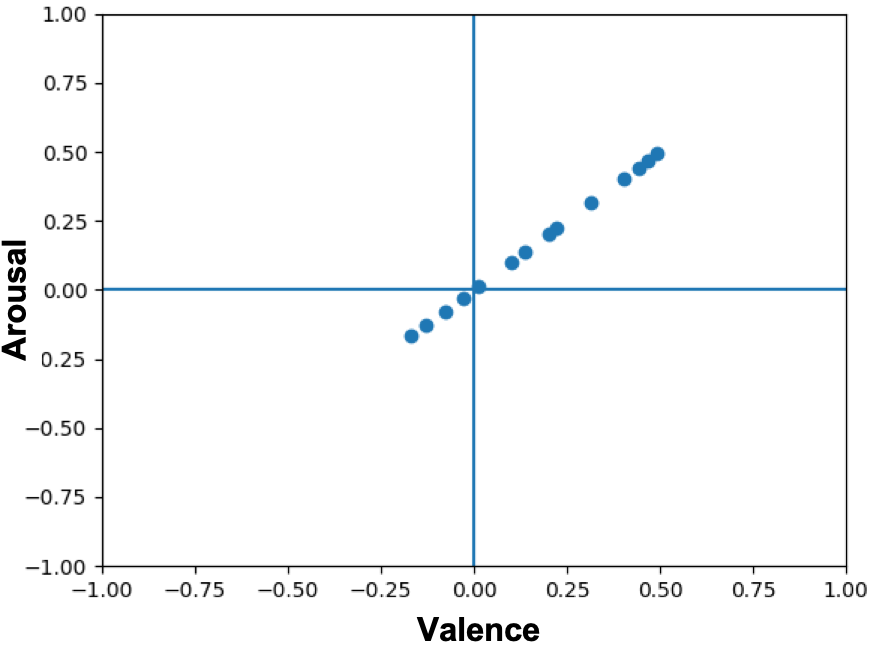}}
\subfloat[Impatient Affective Core.\label{fig:impatient_affcore}]{\includegraphics[width=0.246\textwidth]{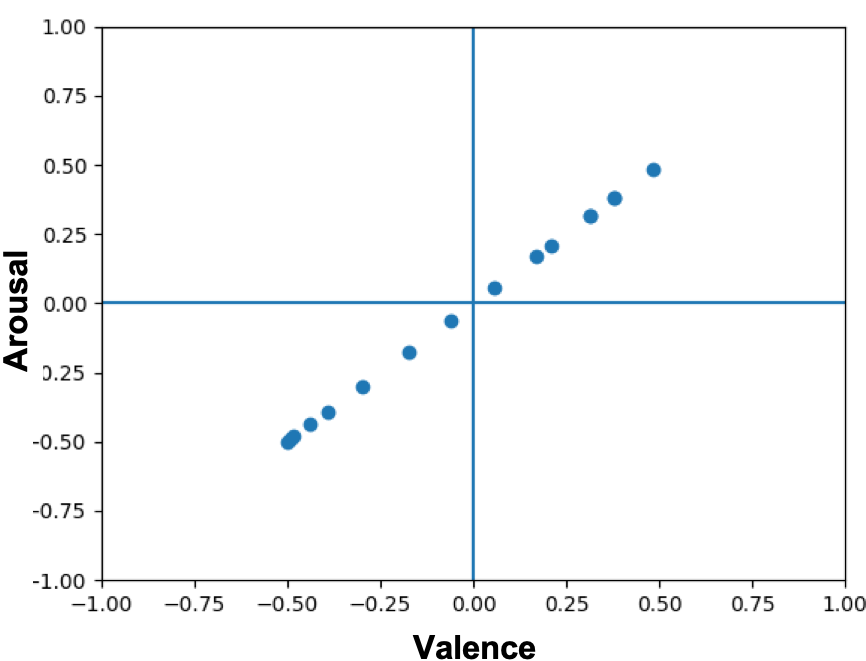}}
\caption{Affective Core biases encoding the impact of Time Perception on the affective appraisal of the agent.}
    \label{fig:time_affcore}
    \vspace{-3mm}
    \end{figure*}

\vspace*{-1mm}
\subsection{Modelling the Affective Core of the Robot}
\label{sec:core}

The \textit{Affective Core} in humans acts as an emotional disposition, not just contributing towards their affective appraisal, but also governing their behaviour~\cite{EMDE1991251}. Similarly, an affective core for a robot can be used as the basis for inherent personality traits that may influence its perception and behaviour. This work explores the influence of two intrinsic qualities, namely \textit{time perception} and the \textit{social conditioning} of the robot, forming the \textit{affective core} of the robot (see Fig.~\ref{fig:mmemotion}). While time perception refers to how the robot is impacted by the duration of an interaction, social conditioning accounts for the \textit{acculturation} or emotional actuation of the robot as a result of its repeated interactions with affective stimuli. These qualities, amongst others, are also found to have an influence on personality formation in infants~\cite{Rothbart2000Temperament} resulting from engagement with caregivers. 

We propose the use of Recurrent Gamma-\ac{GWR} models~\cite{PARISI2017Lifelong}, equipped with a Gamma-context memory~\cite{VRIES1992565}, for modelling the \textit{affective core} of the robot. Yet, rather than focusing on the temporal evolution of an expression, for example, \textit{onset} to \textit{offset} for a facial expression~\cite{Sariyanidi2015Automatic}, we focus on tracking the evolution of the overall affective behaviour over several time-steps. The encoded arousal-valence values obtained by classifying the feature prototypes resulting from the perception model are examined over the entire duration of the interaction. To account for such temporal dynamics, each neuron is equipped with a fixed number of context descriptors which increase the temporal resolution of the model.

The learning rule and activation functions for the \ac{GWR} model~\cite{MARSLAND20021041} are modified to account for activation of the neurons from the previous \bm{$K$} (number of \textit{Gamma} filters) time-steps. The \ac{BMU} or winner neuron $b$ is computed as follows:

{\small
\begin{equation}
b = \underset{i}{\arg \min} \{d_i\},
\label{eq:1}
\end{equation}
\vspace*{-3mm}
}

where $d_i$ is the distance of the neuron $i$ from the data-point. The activation takes into account both the distance between the input and the weights at the current time-step as well as uses the context activation over the last $K$ gamma filters:
{\small
\begin{equation}
d_i = \alpha_{w} \ . \ ||x(t) - w_i ||^2 \ + \ \sum_{k=1}^{K} \ \alpha_{k} \ . \ ||C_{k}(t) - c_{i}^{k}||^2,
\end{equation}
}
where $x(t)$ represents the current input (in this case, the $200$-d combined dense representation for the Perception-\ac{GWR} and the \textit{affective memory} and the encoded $2$-d arousal-valence value for the \textit{affective core}), $w_i$ represents the weight vector of the $i$\textsuperscript{th} neuron, $\alpha_w$ and $\alpha_k$ are constants influencing the modulations from past activation and the current input, $C=[c_{1}^i,c_{2}^i,\dots,c_{k}^i]$ is the set of context vectors for the $i$\textsuperscript{th} neuron with $k=1,2,\dots,K$ being the Gamma filter order. Global context $C_{k}(t)$ is given as:
{\small
\begin{equation}
C_{k}(t) = \beta \ . \ w_{b(t-1)} \ + \ (1-\beta ) \ . \ c_{b(t-1)}^{k-1}
\end{equation}
}
where $\beta$ controls the influence of the previous activation on the current processing of input, $b(t-1)$ is the winner neuron from the previous time-step and $c_{b(t-1)}^0 \equiv w_{b(t-1)}$.

Once the \ac{BMU} is selected, the weight of the winning neuron and the context vectors are updated as follows:
{\small
\begin{equation}
\Delta w_i = \epsilon_i \ . \ \eta_i \ . \ (x(t) - w_i),
\end{equation}
\begin{equation}
\Delta c_{i}^k = \epsilon_i \ . \ \eta_i \ . \ (C_k(t) - c_{i}^k),
\end{equation}
}
where $\epsilon_i$ is the learning rate that modulates the updates and does not decay over time. The firing counter $\eta_i$, on the other hand, is used to modulate learning~\cite{MARSLAND20021041}. It is initialised to $1$ ($\eta_0=1$) and decreased according to the following rule:

{\small
\begin{equation}
\Delta \eta_i = \tau_i \ . \ \kappa \ . \ (1 - \eta_i) - \tau_i
\label{eq:6}
\end{equation}
}
where constants $\kappa$ and $\tau_i$ control decay curve behaviour. 

\subsubsection{Interaction Time Perception}
\label{sec:timePerception}
Starting from the same initial affective state, the robot, given its inherent time perception, maintains its intrinsic state for the entire duration of the interaction with a user. To simulate the impact of time, a decay function ($y=\exp(-\tau t)$) is implemented that modulates the affective state of the robot at any given time. For simulating \textit{patience}, the decay is slow and gradual ($\tau=0.01$) allowing the robot to maintain its affective state for a longer duration while in case of \textit{impatient} time perception, this decay is rapid ($\tau=0.08$). The decay function dynamics can be seen in Fig.~\ref{fig:patient_time} and Fig.~\ref{fig:impatient_time}, respectively. The empirical choice of $\tau$ values assures smooth decay curves over a minimum of $90$ time-steps. These values can be adapted as per the desired impact of time perception in the robot.


Given a \textit{patient} or \textit{impatient} modulation, an \textit{affective core} bias is modelled using the Gamma-\ac{GWR} model. The initial state of the robot is set to a mean positively excited arousal, valence values of $(0.5,0.5)$ and then modulated over time using the decay function $y$. At each time-step, the Gamma-\ac{GWR} model receives this modulated input state and forms intrinsic prototypes following the process described in Eq.~\ref{eq:1}$-$\ref{eq:6}. This models the decay dynamics of the robot's intrinsic state at different time-steps, forming a time perception bias that encodes \textit{patient} (Fig.~\ref{fig:patient_affcore}) or \textit{impatient} (Fig.~\ref{fig:impatient_affcore}) behaviour.

\begin{figure}
\vspace{-3.5mm}

    \centering
    \subfloat[Excitatory Affective Core.\label{fig:high_core}]{\includegraphics[width=0.25\textwidth]{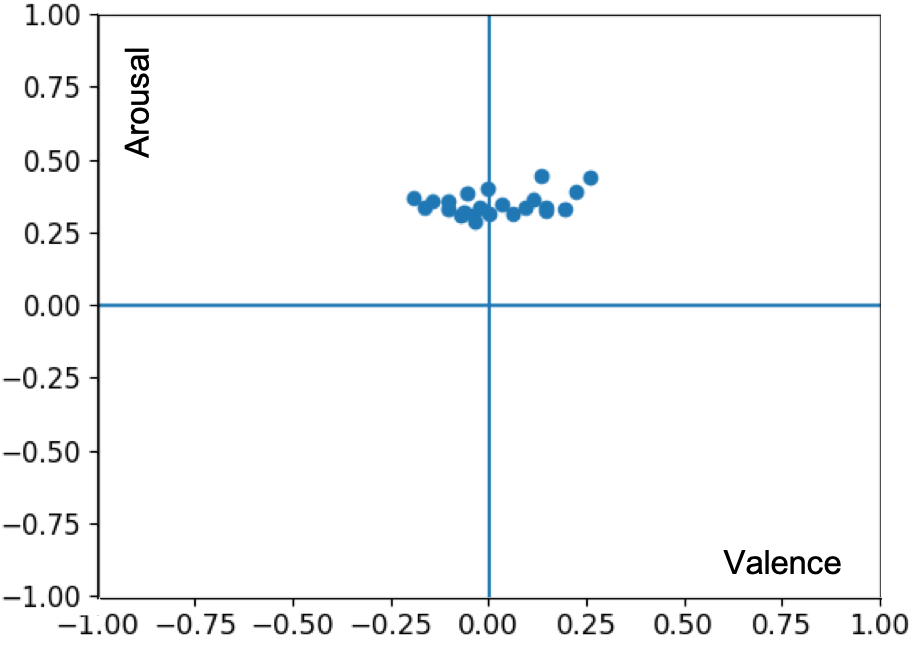}}
    \subfloat[Inhibitory Affective Core.\label{fig:low_core}]{\includegraphics[width=0.25\textwidth]{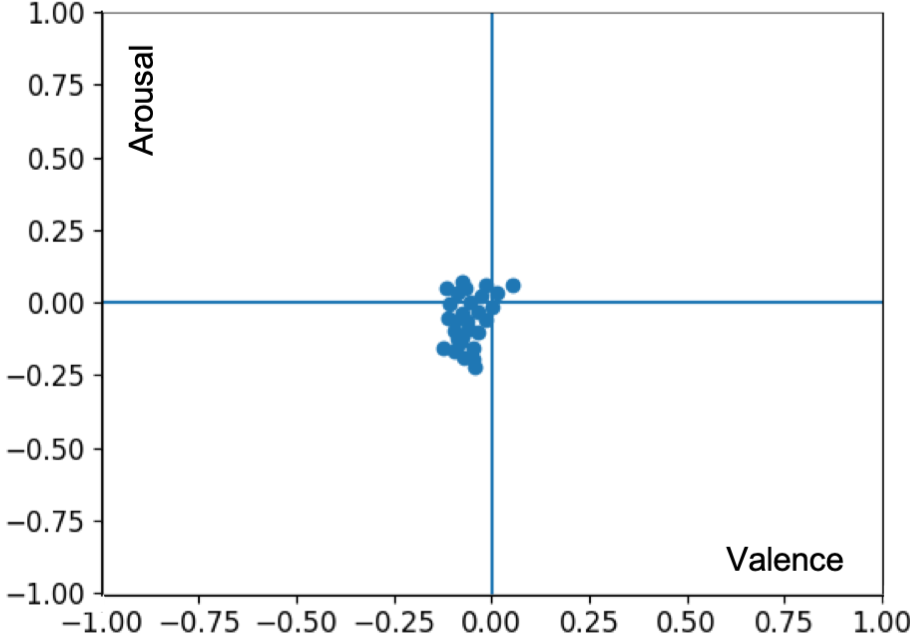}}
\caption{Affective Cores modelling high or low arousal perception input to form different social conditioning biases.}
\label{fig:socialcores}
\vspace{-3.5mm}
\end{figure}

\subsubsection{Social Conditioning}
\label{subsubsec:Social}

Social conditioning of the robot can be used to formulate anchors for its affective appraisal. The robot, through continued and repeated interaction with affective stimuli, can get acculturated, developing qualities central to its \textit{personality}. Such conditioning can be \textit{excitatory} (high-arousal), amplifying the impact of perception, or \textit{inhibitory} (low-arousal), diminishing it. To model such influences, the robot is shown videos (see Section~\ref{subsec:MoodAffCore} for details) encoding different emotional intensities. To model an excitatory effect, videos encoding high arousal are used whereas, for the inhibitory core, low arousal videos are used. The videos are processed using the \ac{MCCNN} - Perception-\ac{GWR} model. \ac{BMU}s from the Perception-\ac{GWR}, for every $500$ milliseconds of audio-visual input, are classified into the corresponding arousal (A) and valence (V) values they encode and used as input for the \textit{affective core} Gamma-\ac{GWR} model. Training parameters used for the model can be seen in Table~\ref{tab:parameters}. The resultant, prototypical,  \textit{excitatory} core (Fig.~\ref{fig:high_core}) represents only high-arousal information (\bm{$A$}$>0.3$) while the \textit{inhibitory} core (Fig.~\ref{fig:low_core}) encodes low-arousal information (\bm{$A$}$<0.05$).

\vspace*{-1.7mm}
\subsection{Mood Formation for the Robot}
\label{sec:mood}
The intrinsic \textit{mood} forms the affective appraisal of a robot, estimating its intrinsic state during interactions. Given past experiences with a user, the robot monitors their affective behaviour to conclude their affective state. This input, modulated by the robot's affective core, elicits an emotional response in the robot in the form of its \textit{mood} (see Fig.~\ref{fig:mmemotion}), acting as the motivation for subsequent interactions.

In this work, we model robot's mood as a Gamma-\ac{GWR}~\cite{PARISI2017Lifelong} (following Eq.~\ref{eq:1}$-$\ref{eq:6}) that evaluates the current behaviour of the user (\textit{affect perception}), modulated by past experiences (\textit{affective memory}) to form an intrinsic affective response in the robot towards the user. This is further influenced by the robot's \textit{affective core}. All these inputs are processed \textit{asynchronously} to allow for the evolution of the robot's mood even when they are sparsely available. This results in the robot forming an organic affective response towards the user rather than merely mimicking them. Different \textit{affective core} biases in the robot result in the same stimulus being evaluated differently. For example, a \textit{patient} robot with an \textit{excitatory} conditioning will be able to retain its positive mood for longer, despite receiving a series of negative inputs. This is important as it can be used to integrate different personality traits in the robot, with different combinations of the affective core influences (see Table~\ref{tab:result_affcore_bias}) yielding significantly different \textit{mood} estimates. 

\vspace*{-1mm}
\subsection{Learning Robot Behaviour}
\label{sec:behavemodel}
The affective appraisal of the robot, under specific \textit{affective core} traits can help learn generate different robot behaviours in collaborative \ac{HRI} scenarios. In this work, we explore the Ultimatum Game~\cite{Guth1982An} to embed different negotiating behaviours in the robot, given its affective core. We propose, a \acf{DDPG}-based actor-critic model~\cite{Lillicrap2015Continuous} that learns to interact with human participants, incorporating the robot's \textit{mood}, both in the state-value function as well as in the reward received by the robot. The proposed model aims to evaluate how the robot, given its personality traits, can learn to successfully negotiate resources with human participants. Furthermore, evaluation of the robot by the participants under different \textit{affective core} conditions can highlight the contribution of the robot's personality towards its negotiation capabilities.

\begin{figure}
    \centering
    \subfloat[Participant rejects the offer.
  \label{fig:part-reject}]{\includegraphics[width=0.24\textwidth]{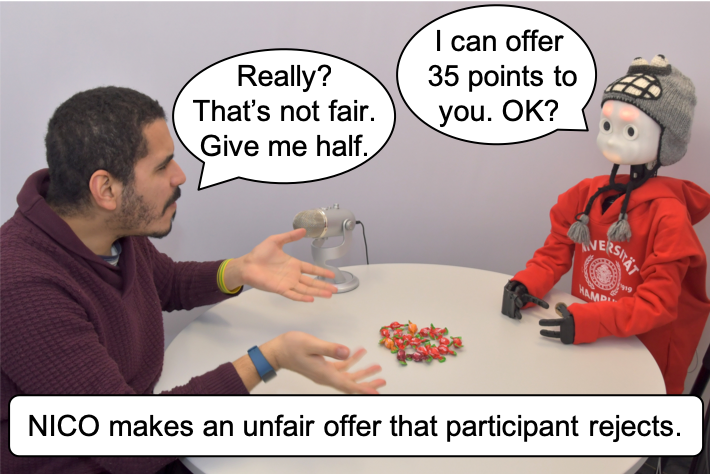}} \hspace{0.01mm}
    \subfloat[Participant accepts new offer.
  \label{fig:part-agree}]{\includegraphics[width=0.24\textwidth]{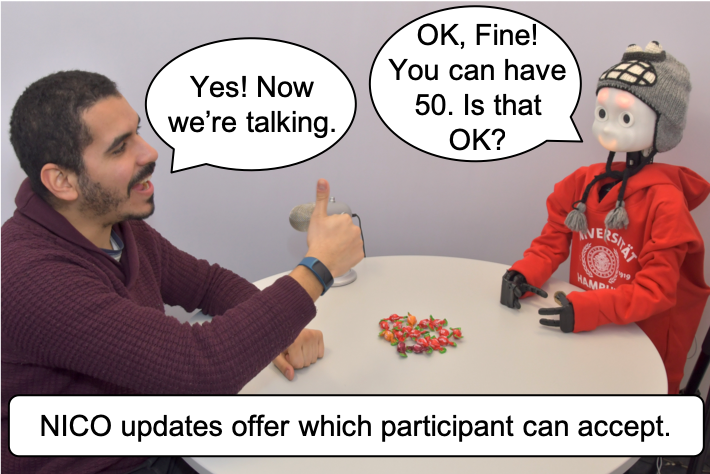}} 
    \caption{Participant and \acs{NICO} in the Ultimatum Game scenario.}
    \label{fig:negotiate}
    \vspace{-3mm}
\end{figure}

\subsubsection{The Ultimatum Game}
\label{subsec:UG}
The traditional design for the Ultimatum Game~\cite{Guth1982An} involves two participants namely, a \textit{proposer} and a \textit{respondent}, negotiating a split of resources (usually money). The \textit{proposer} offers a split, based on which the \textit{respondent} either \textit{accepts} or \textit{rejects} the offer. Only if the offer is accepted, resources are shared as per the agreed split.

We extend this design by incorporating a negotiation between the participants (see Fig.~\ref{fig:negotiate}) and the \acf{NICO} robot~\cite{Kerzel2017NICO} acting as the \textit{proposer}. \ac{NICO} and each participant are given $100$ points that can be exchanged for $20$ \textit{bonbons}, with every $5$ points fetching them one bonbon. Bonbons are used as they give a visual motivation for the negotiation. As \ac{NICO} makes offers to the respondent, if they accept the offer, the interaction culminates with both receiving the agreed split. In case of a rejection, \ac{NICO} asks the participants for reasons for their rejection and based on their affective responses, it appraises their affective state as an evaluation of the offer, eliciting a change in its own mood. This mood (or change in the mood thereof) is used to update its offer in a way that the participant may accept the new offer, without \ac{NICO} losing a lot of points. The participant and \ac{NICO} thus negotiate a split of the $100$ points with \ac{NICO} updating its offer upon each rejection. To assure that negotiations come to a conclusion, the negotiation is aborted with no one getting any points after the participant rejects $20$ consecutive offers.

\subsubsection{Learning to Negotiate}
\label{subsec:behavelearn}

While negotiating with the participants, the intrinsic \textit{mood} of the robot after each rejection, concatenated with the rejected offer value, is mapped to the state-space of the robot to generate actions in the form of increments or decrements on the previous offer. This results in a continuous, high-dimensional action-space making the use of traditional $Q$-learning approaches difficult as they become intractable in such high-dimensional spaces~\cite{Lillicrap2015Continuous}. Also, it is desirable that these updates to the offer are not modelled as fixed increments or decrements to enable a more naturalistic negotiation between participants and \ac{NICO}. Thus, this work employs a \acs{DDPG}-based actor-critic model~\cite{Lillicrap2015Continuous} to learn an optimal negotiating behaviour. 

\begin{figure}
    \centering
    \includegraphics[width=0.438\textwidth]{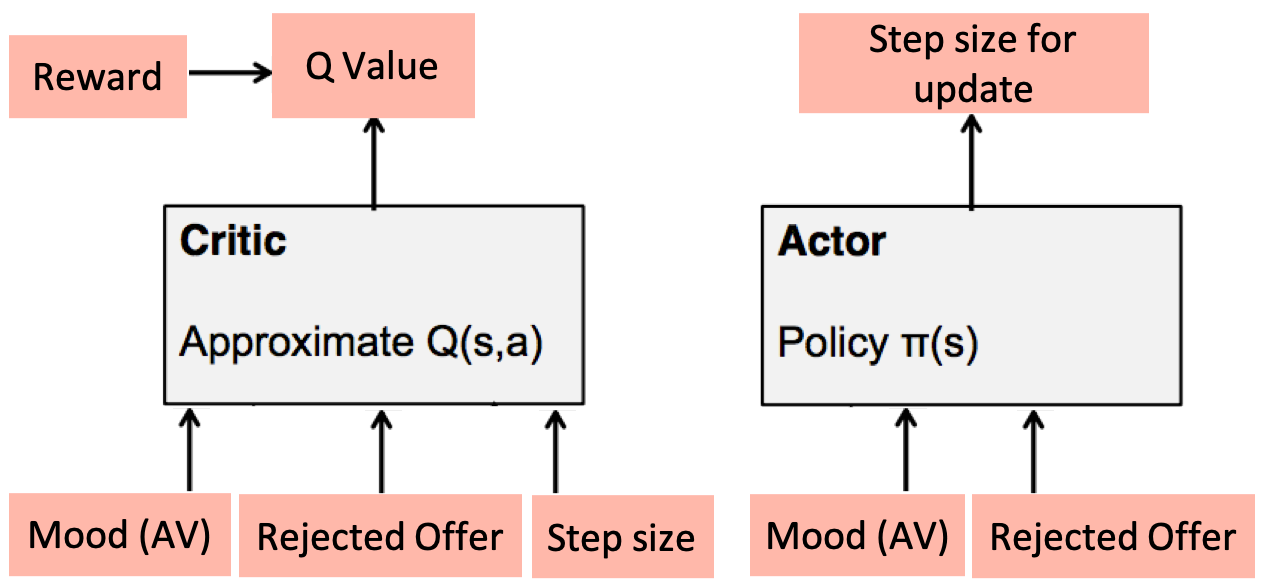}
    \caption{Actor-Critic model for Learning Robot behaviour.}
    \label{fig:behaveModel}
    \vspace{-3.5mm}
\end{figure}

The model consists of two separate models (see Fig.~\ref{fig:behaveModel}) for the \textit{actor} and the \textit{critic}, respectively. The actor takes the robot's current mood (mean arousal-valence vector computed from all the neurons of the mood Gamma-\ac{GWR} model (see Section~\ref{sec:mood})) as well as the previously rejected offer\footnote{$(r,h)$-vector where $r$ is the robot's share and $h$ is the participant's share} as inputs and concatenates them into a single $4$-tuple representing the state of the robot. This state is passed to the \textit{actor}, predicting a real-valued update over the previous offer. 

The \textit{critic} network takes the current state of the robot as well as the actor-generated update value as inputs to evaluate the actor's performance, predicting a $Q$-value~$\in\mathbb{R}$ for the state-action pair. This predicted $Q$-value and the reward received by the robot (see Section~\ref{sec:offlineRobot} for details) are used to update both the critic and the actor~\cite{Lillicrap2015Continuous}. 

The actor and critic are modelled as \ac{MLP} networks (see Fig.~\ref{fig:mmemotion}). The actor network takes as input the current state of the robot ($4-$tuple). This input is connected to a FC layer consisting of $50$ units which is further connected to an output neuron, predicting real-valued updates on the offer. For the critic, the state $4-$tuple and the predicted update value are connected to individual FC layers of $50$ units each. These FC layers are then concatenated and connected to another FC layer of $10$ units combining the representations. Finally, a single output neuron predicts the $Q$-value~$\in\mathcal{R}$ for the state-action pair.

\vspace*{-1mm}
\section{Experiments and Results}
\subsection{Multi-Modal Affect Perception} 
\subsubsection{The \ac{MCCNN} Network}
\label{sec:train_MCCNN}
Training the \ac{MCCNN} model requires multi-modal datasets that provide good quality samples for both vision and speech modalities with continuous arousal-valence annotations. Most of the available multi-modal datasets rely on the visual information as the dominant modality deciding affective labels. This is seen in the \acs{Aff-Wild}~\cite{Zafeiriou2017AFF} and \acs{AFEW-VA}~\cite{KOSSAIFI201723} datasets where the audio samples are affected by background music or noise. On the other hand, datasets like \acs{RAVDESS}~\cite{livingstone2012ravdess} and \acs{SAVEE}~\cite{HaqJackson_MachineAudition10} provide clean audio and video samples but use categorical labelling. 

Thus, we pre-train the face-channel of \ac{MCCNN} combining \acs{Aff-Wild} and \acs{AFEW-VA} datasets with normalised arousal-valence labels $\in[-1,1]$ for each frame. The face-channel is trained with a $60$:$20$:$20$ (train, validation, test) data split reaching competitive \ac{CCC} scores of \bm{$0.68$} for \textit{arousal} and \bm{$0.57$} for \textit{valence} (compared to baselines~\cite{Zafeiriou2017AFF, KOSSAIFI201723}). The face-channel model is then used to classify facial images from the \acs{RAVDESS} and \acs{SAVEE} datasets, generating arousal and valence labels. These labels are then used to train the combined \ac{MCCNN} network using audio-visual information. This approach is inspired from Lakomkin~et~al.~\cite{LWW17} who conclude that augmenting datasets using labels from one modality contributes positively towards improving the overall performance of the model. The \ac{MCCNN} is trained for $200$ epochs with the \textit{Adam} optimiser on the combined dataset (\acs{RAVDESS} and \acs{SAVEE}), converging to \textbf{\ac{CCC}} scores of \bm{$0.75$} for \textit{arousal} and \bm{$0.53$} for \textit{valence}. The hyper-parameters for \ac{MCCNN} are optimised using the Hyperopt\footnote{\url{https://github.com/hyperopt/hyperopt}} python library. 


\begin{table}
\centering
\footnotesize{
\caption{Training parameters for the Self-Organising Models.}
\label{tab:parameters}\vspace{-1mm}
\setlength\tabcolsep{2.5pt}
\begin{tabular}{cc|c|c|c}
\toprule
\textbf{Model}  &  \makecell{\textbf{Habituation}\\\textbf{Threshold}}   & \makecell{\textbf{Insertion}\\\textbf{Threshold}} &  \makecell{\textbf{Max.}\\\textbf{Age}}   &   \makecell{\textbf{Context}\\\textbf{Vectors}} 
\\\midrule
\makecell{Perception-GWR}            &   0.2  &   0.5   &  50 &   --\\
\makecell{Affective Memory}          &   0.5  &   0.8   &   5 &   10\\
\makecell{Time Perception Core}      &   0.5  &   0.9   &   5 &   5\\
\makecell{Social Conditioning Core}  &   0.5  &   0.9   &   5 &   5\\
\makecell{Robot Mood}                &  0.5  &   0.9   &   5 &   10\\\bottomrule
\end{tabular}
}
\vspace{-5mm}
\end{table}

\subsubsection{Perception-\ac{GWR} and the Affective Memory}
\label{sec:train_GWR}
For training the Perception-\ac{GWR} model, feature vectors from the $200$-d FC layer of the \ac{MCCNN} are extracted for the entire training set (combined \acs{RAVDESS} and \acs{SAVEE}). The \ac{GWR} model is trained for $50$ epochs with a maximum age of $50$ for each neuron to allow for a neuron to be retained even if it fires only once per epoch. The habituation threshold (see Table~\ref{tab:parameters} for details) controls the frequency of weight updates while the insertion threshold controls when a new neuron needs to be added. This results in a total of \bm{$458$} \textbf{neurons} which sufficiently represent the entire training set ($\approx20$k data points). These neurons act as feature prototypes for the entire dataset, enabling a robust evaluation of the arousal-valence represented in the data samples. Fig.~\ref{fig:mmemotion} shows the Perception-\ac{GWR} with each neuron plotted according to the arousal and valence it encodes. The choice of the different thresholds is determined empirically, given the resultant \ac{GWR}'s ability to represent the training set.

The \textit{affective memory} Gamma-\ac{GWR} model consists of $10$ context descriptors, implementing a temporal resolution of $10$ time-steps. The model is trained following Eq.~\ref{eq:1}$-$\ref{eq:6}. The chosen parameters (see Table~\ref{tab:parameters}) allow the model to map, and remember, the affective context for at least one complete interaction ($5-8$ seconds). Fig.~\ref{fig:mmemotion} shows the affective memory for the user with each neuron plotted according to the arousal and valence it encodes. The insertion and habituation thresholds control the update of existing neurons and add new neurons only when needed. A separate \textit{affective memory} is created for each user interacting with the robot.

\vspace*{-1mm}
\subsection{Mood Formation under Affective Core Influence}
\label{subsec:MoodAffCore}
To evaluate the impact of the different \textit{affective core} influences on the mood formation of the robot, $20$ videos each from the \acs{KT} Emotion Dataset~\cite{Barros2017ASelf} and the \acs{OMG-Emotion} Dataset~\cite{Barros2018OMG} are selected as both these datasets consist of clean audio-visual samples encoding different affective contexts. Each video is split into data-chunks representing $500$ milliseconds of audio-visual information. The pre-trained Face Detector from the Dlib python library is used for extracting faces while the python Scipy Signal processing library is used to generate mel-spectrograms for each data-chunk. The data is input sequentially to the \ac{MCCNN} and perception-\ac{GWR} for feature extraction and representation while the different Gamma-\ac{GWR} networks model the \textit{affective memory} and \textit{affective core} biases.

The mood Gamma-\ac{GWR} model is trained for $10$ epochs taking as input, for every $500$ milliseconds of audiovisual input, $2$ \ac{BMU}s from the Perception-\ac{GWR} encoded into the arousal-valence values they represent, the mean arousal-valence vector from the \textit{affective memory}, $5$ \ac{BMU}s from the social conditioning Gamma-\ac{GWR} and $2$ \ac{BMU}s from the time perception Gamma-\ac{GWR}. Different combinations of \textit{affective core} biases are explored to evaluate how these influence mood formation in the robot. A Two-Sided Mann-Whitney~U test shows significant differences (see Table~\ref{tab:result_affcore_bias}) in the resultant mood under different affective cores, compared to when no \textit{affective core} bias is used. The model is shown the same video sequences changing only the \textit{affective core} between repetitions. Keeping all other variables constant, any change in the resultant mood can be attributed to the \textit{affective core}.

\begin{figure}[t]
\centering
\includegraphics[width=0.5\textwidth]{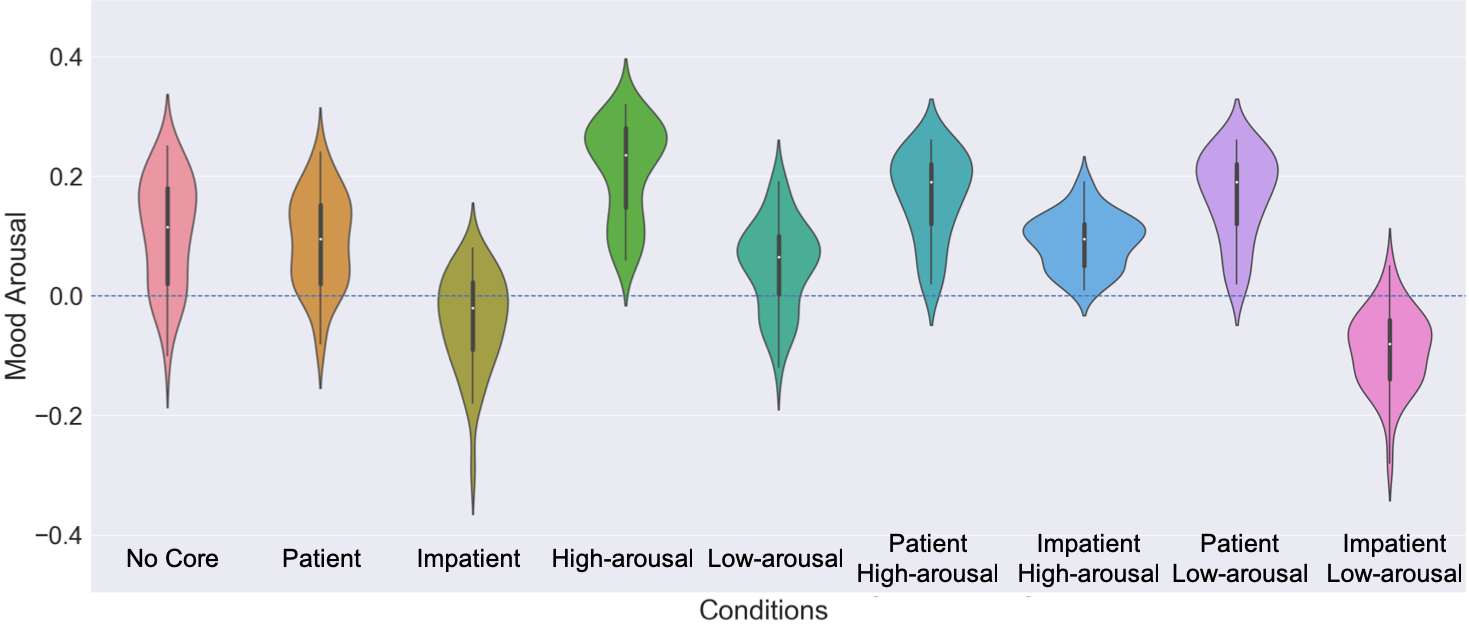}
\caption{Arousal distribution for different affective core biases.}
\label{fig:ar-box}
\vspace{-3mm}
\end{figure}

\begin{figure}[t]
\centering
\includegraphics[width=0.49\textwidth]{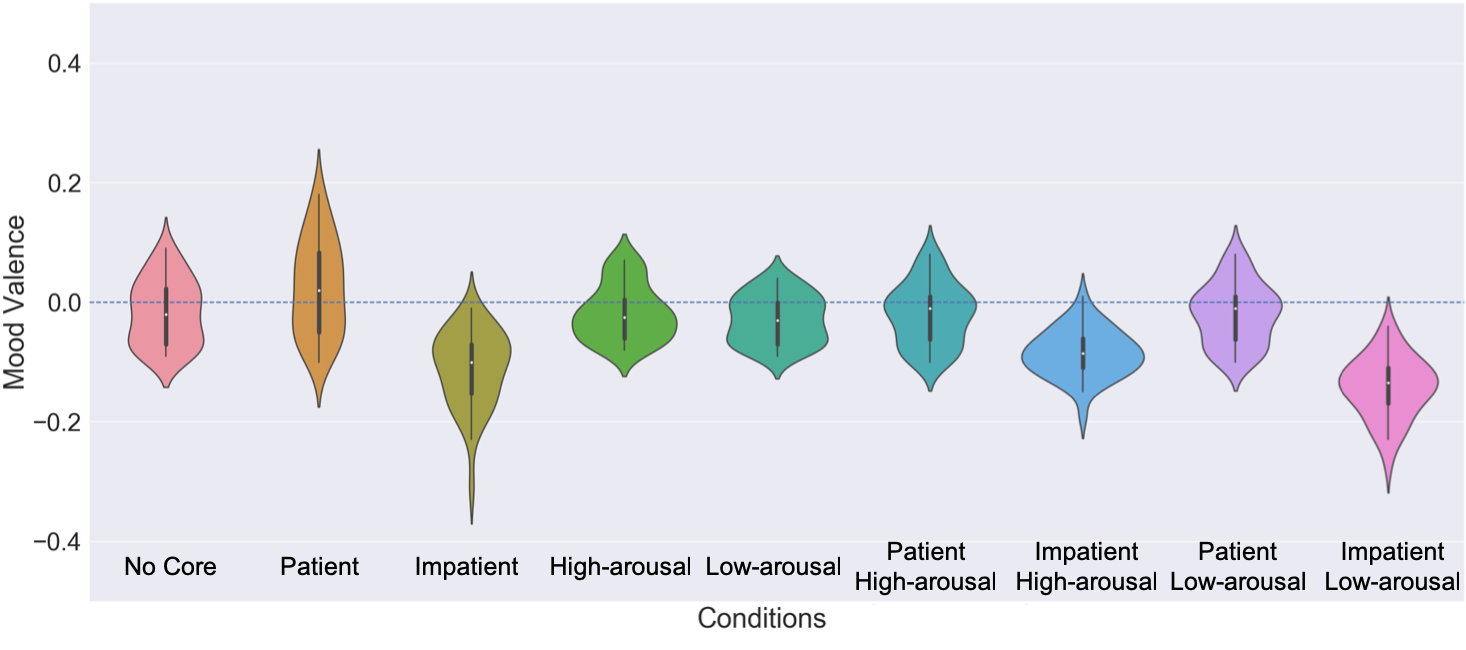}
\caption{Valence distribution for different affective core biases.}
\label{fig:val-box}
\vspace{-3mm}
\end{figure}

\begin{table}[t] 
\centering
\caption{Two-sided Mann-Whitney U-test results with the alternative hypothesis that the resultant mood under different affective cores is different from the \textit{No Core} condition.}
\label{tab:result_affcore_bias}
\vspace{-1mm}

\footnotesize{
\setlength\tabcolsep{1.0pt}
\hspace*{-1mm}\begin{tabular}{ c c c c c c c} 
\toprule
\makecell{\rule{0pt}{2ex} \textbf{Time}\\\textbf{Perception}}  & \makecell{\textbf{Social}\\\textbf{Conditioning}} &
\makecell{\textbf{Affective}\\\textbf{Core}} & \makecell{\textit{\textbf{U-statistic}}\\\scriptsize{\textit{Arousal}}} & \makecell{\textit{\textbf{\textit{p-value}}}\\\scriptsize{\textit{Arousal}}} & \makecell{\textit{\textbf{U-statistic}}\\\scriptsize{\textit{Valence}}} & \makecell{\textit{\textbf{\textit{p-value}}}\\\scriptsize{\textit{Valence}}}\\ \midrule
None & Excitatory & High-arousal 			& \textbf{317.0} 		& \textbf{$<<$0.05}   & 793.5 			& 0.9520  			 \\ \midrule
None & Inhibitory & Low-arousal 			& \textbf{500.5} 		& \textbf{$<<$0.05}		& 709.0 			& 0.3840 			 \\ \midrule
Patient & None & Patient 				& 703.0 				& 0.352				& \textbf{559.0} 	& \textbf{0.02}	 \\ \midrule
Impatient & None & Impatient 				& \textbf{186.0} 		& \textbf{$<<$0.05}	& \textbf{211.0} 	& \textbf{$<<$0.05}\\\midrule
Patient & Excitatory & \makecell{Patient\\High-arousal} 	& \textbf{488.0} 		& \textbf{$<<$0.05} 		& 766.5 			& 0.7480 			 \\ \midrule
Patient & Inhibitory & \makecell{Patient\\Low-arousal} 	& \textbf{412.5} 		& \textbf{$<<$0.05} 		& 694.5 			& 0.3120 			 \\ \midrule
Impatient & Excitatory & \makecell{Impatient\\High-arousal} 	& 674.5 				& 0.2300 				& \textbf{251.0} 	& \textbf{$<<$0.05}\\  \midrule
Impatient & Inhibitory & \makecell{Impatient\\Low-arousal} 	& \textbf{57.5} 		& \textbf{$<<$0.05} 	& \textbf{52.5} 	& \textbf{$<<$0.05}\\ \bottomrule
\end{tabular}
}
\vspace{-4mm}

\end{table}
The model, for the same input stimuli, is seen to form different estimates of its intrinsic mood under different \textit{affective core} biases (see Fig.~\ref{fig:ar-box} for arousal and Fig.~\ref{fig:val-box} for valence distributions). The affective appraisal of the robot under the different \textit{affective core} biases is compared to the \textit{No Core} condition which considers only the agent's current perception and its \textit{affective memory} for mood formation. 

The arousal values show more deviation from the baseline due to the \textit{excitatory} or \textit{inhibitory} effect of the \textit{affective core}. Since these biases predominantly affect the intensity of the robot's intrinsic mood, the corresponding plots for the valence show much less deviation. On the other hand, the \textit{patient} or \textit{impatient} biases impact both the valence and arousal.

This effect is validated by conducting a Two-sided Mann-Whitney~U test~\cite{Mann1947} using the resultant mood estimates (arousal-valence) under different \textit{affective core} biases with the alternative hypothesis that the resultant mood is different from the \textit{No Core} condition. Robot mood results in significantly different arousal and valence distributions (see Table.~\ref{tab:result_affcore_bias}) for the same input. Even though the input videos are chosen to cover different affective contexts, the intrinsic mood of the agent is influenced by the respective \textit{affective core}. For a social robot, this means that rather than mimicking the user, the robot, true to its intrinsic personality traits, can formulate distinct affective responses towards the user. 


\vspace*{-1mm}
\subsection{Pre-Training Robot Behaviour Model Off-line}
\label{sec:offlineRobot}
The intrinsic mood of the robot is used as the motivation to learn different robot behaviours in the Ultimatum Game. As a negotiation might be short-lived and may not provide with sufficient examples to train an \ac{RL} model, we pre-train the model using a probability-based reward function. The acceptance of an offer is modelled in a stochastic manner (see Eq.~\ref{equation}) based on the fraction of the resources being offered to the participant. 

\vspace*{-1mm}

{\footnotesize \begin{equation}
p(\text{acceptance}) = 
\begin{cases}
    1,				        & \text{if } \text{offer } \geq 0.7\\
    \text{\emph{offer}},	& \text{if } 0.7 > \text{offer } \geq 0.5\\
    0.1,		            & \text{if } 0.5 > \text{offer } \geq 0.4\\
    0,                      & \text{otherwise}
\end{cases}
\label{equation}
\end{equation}
}
\vspace*{-1mm}

The reward function models two competing goals for the robot; keeping a higher share of resources for itself and, at the same time, eliciting a positive response from the respondent. The two components of the reward are explained as:
\begin{itemize}[leftmargin=*]
    \item The \textit{offer reward} provides intermediate positive rewards if the new offer increases the respondent's share while keeping the robot's share $>50\%$. These rewards smoothen the learning curve, guiding the robot to an optimal behaviour.
    \item The \textit{mood reward} computes a (cosine) distance measure between the previous and the new mood state of the robot and rewards a positive change in the robot's mood. As the robot's mood reflects the respondent's affective state, it learns to evoke positive responses to its offer.
\end{itemize}  

The \ac{DDPG} model is pre-trained using data samples generated by processing $20$ videos from the \acs{KT} Emotion dataset through the \ac{MCCNN} - Perception-\ac{GWR} model and selecting the two \ac{BMU}s. This is augmented by adding randomly generated arousal-valence vectors to cover the entire state-space. A total of $500$ random samples are added, drawn from a standard normal distribution sliced to range~$\in [-1,1]$. To match video dynamics, each added sample undergoes an interaction decay (forming a trajectory) simulating robot mood. This decay emulates affective responses from a respondent that witnesses consecutive unfair offers from the robot and rejects them. 

\begin{figure}
    \centering
    \subfloat[Average number of negotiations.\label{fig:learningnegotiations}]{\includegraphics[width=0.235\textwidth]{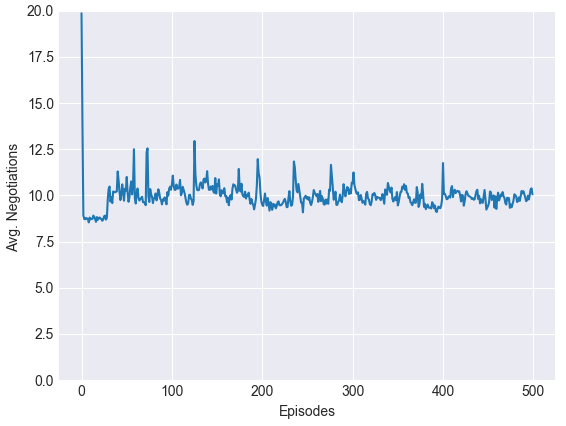}} \hspace{0.1em}
    \subfloat[Average reward per episode.\label{fig:avgreward}]{\includegraphics[width=0.23\textwidth]{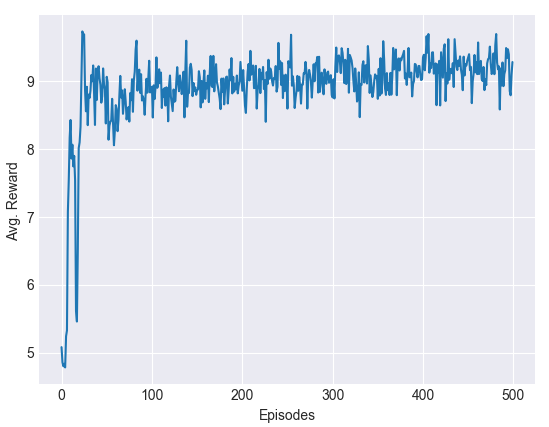}}
    \caption{Robot learning to negotiate, converging on an acceptable offer $>45\%$ of the resources in $\approx10$ interactions.}
    \label{fig:robotnego}
    \vspace{-3mm}
\end{figure}

The model balances both the offer and mood reward and converges to offering $40-60\%$ points, yielding an optimal reward for the robot (see Fig.~\ref{fig:avgreward}). The average number of interactions reduce to~$\approx10$ as the model learns to find an optimal offer that faces fewer rejections (see Fig.~\ref{fig:learningnegotiations}). 

\vspace*{-1mm}
\subsection{User Study}
\label{sec:behaveexp}

To investigate whether the \textit{affective core} of the robot results in different negotiation strategies in the Ultimatum Game, we conducted a user study that assessed how different participants evaluate \ac{NICO}'s behaviour (realised using the pre-trained \ac{RL} model). Furthermore, quantitative performance factors like success-rate (acceptance of the robot's offer), mean accepted offer value, and the number of interactions are also evaluated.

The user study was conducted with $31$ participants ($20$ male, $11$ female) from $16$ countries in the age group of $18-49$. All participants, recruited amongst university students and employees, reported conversational proficiency in English (the language used to model interactions). The participants were briefed about the objectives of the experiment and the interaction procedure and they provided \textit{informed consent} for their participation. The consent form and the experiment protocol were approved by the Ethics Commission\footnote{\url{https://www.inf.uni-hamburg.de/en/home/ethics.html}} of the Department of Informatics, University of Hamburg.

The experiment set-up (see Fig.~\ref{fig:negotiate}) consists of an artificially well-lit room to exclude effects of changing natural lighting conditions. The participants and \ac{NICO} are positioned across a round-table, opposite to each other. \textit{Bonbons} are placed on the table along with a microphone.

\subsubsection{Experiment Conditions}
\label{subsec:conditions}
The user study is conducted as a \textit{between-group} study with two condition groups. Each group consists of two sub-conditions implementing the \textit{No Core} condition as the \textit{baseline}, along with one of the \textit{measured} conditions. In the \textit{No Core} condition, the robot is not embedded with any \textit{affective core} and considers only the perception input for its intrinsic mood. The two condition groups are: 
\begin{itemize}[leftmargin=*]
    \item \textbf{Patient High-arousal}: In this group, the \textit{measured} condition involves the robot with \textit{patient} time perception and \textit{excitatory} social conditioning biases to influence its mood formation. A total of $16$ participants were randomly assigned to this condition group. 
    \item \textbf{Impatient Low-arousal}: In this group, the \textit{measured} condition involves the robot embedded with an \textit{impatient} time perception and \textit{inhibitory} social conditioning bias that influence mood formation. The second condition group consisted of $15$ randomly assigned participants.
\end{itemize}

\subsubsection{Experiment Protocol}
Once the participants are assigned to a condition group, they are introduced to the experiment set-up where \ac{NICO} greets them by modelling a short interaction with the participants informing them about the rules of the game. Google Text-To-Speech python library is used to generate \ac{NICO}'s voice. During this interaction, \ac{NICO} asks the participants about their excitement towards participating in the experiment. With this, it builds a model of its \textit{affective memory} and intrinsic \textit{mood} as a starting point for both sub-conditions. After the introduction round, \ac{NICO} starts negotiating with the participant, randomly loading the first sub-condition. The negotiations consist of two distinct phases:


\begin{itemize}[leftmargin=*]
    \item \textbf{Offer Phase:} \ac{NICO} makes an offer to the participants which they can accept (saying 'Yes') or reject (saying 'No'). A rejection results in \ac{NICO} asking them to explain their rejection while monitoring their affective responses as they describe their opinion about the offer.
    
    \item \textbf{Update Phase:} Observing the participants' responses, \ac{NICO} models its \textit{mood}, as an evaluation. The mood represents the current state of the robot and is used to compute a new offer for the participants.
\end{itemize}

Each condition begins a random unfair offer with at the most $20$ points offered to the respondent to make sure that the participants are inclined to negotiate \textit{at least once} with the robot. Negotiations continue until the participant either accepts the offer or rejects $20$ consecutive times (empirically defined limit). The participants are told that the robot shall abort the negotiation if a \textit{stalemate} is reached, blinding them from this limit to avoid any behavioural conditioning.


After each sub-condition, the participants fill out a \textit{pseudonymised} $3$-part questionnaire about their experience with the robot to measure any reported difference in robot behaviour between the two sub-conditions. Finally, participants are debriefed and informed about the condition they were assigned to. In the absence of any monetary compensation, as a reward for participation, they are offered all the bonbons.


\subsubsection{Quantitative Results}
\label{sec:behaveresults_quan}

For a quantitative evaluation (see Table~\ref{tab:quant}) of the robot's performance under different conditions, several factors are examined. The \textit{success rate} denotes the fraction of participants that accepted the robot's offer. The \textit{average number of interactions} denotes the number of rejections, on average, before an offer was accepted while the \textit{average accepted offer} highlights the accepted offer. The fraction of offers where the participants were offered $50\%$ or more of the points by the robot is also reported.

\begin{table}
\centering
\caption{Quantitative Analysis of \ac{NICO}'s performance in the Ultimatum Game under different experimental conditions.}

\footnotesize{
\begin{adjustbox}{center}
\setlength\tabcolsep{2.0pt}
\begin{tabular}{@{}ccccccc@{}}
\toprule
\multirow{2}{*}{\backslashbox[25mm]{\textbf{Parameter}}{\textbf{Condition}}}  & \multicolumn{2}{c}{\textbf{Baseline}}                  & \multicolumn{2}{c}{\makecell{\textbf{Patient} \\\textbf{High-arousal}}}               & \multicolumn{2}{c}{\makecell{\textbf{Impatient} \\\textbf{Low-arousal}}}    \\\cmidrule{2-7} 
	& \textbf{Mean}         & \textbf{Std. Err}    & \textbf{Mean}   & \textbf{Std. Err}    	& \textbf{Mean} & \textbf{Std. Err}  \\ \midrule
\makecell[l]{Interactions}  & 8.71                & 0.64            & 9.35                  & 1.13     		& 8.33      & 1.20  \\ 
\makecell[l]{First Offer}   & 15                & 0.9           & 14                  & 1.2         & 15      & 1.0  \\ 
\makecell[l]{Accepted Offer}& 44                & 2.0            & 45                  & 1.6         & 43 	    & 2.0  \\ 
\makecell[l]{Final Offer (If rejected)} & 49   & 0.3           & 47                  & 0.9         & 50        & 0.3 \\
\makecell[l]{Offered $>=50$\% }         & \multicolumn{2}{c}{77\%}              & \multicolumn{2}{c}{62\%}              & \multicolumn{2}{c}{80\%}\\ 
\makecell[l]{Success Rate}  & \multicolumn{2}{c}{90\%}              & \multicolumn{2}{c}{87\%}              & \multicolumn{2}{c}{80\%}\\ \bottomrule

\end{tabular}
\end{adjustbox}
}

\label{tab:quant}
\vspace{-4mm}
\end{table}

The \textit{Patient High-arousal} condition, on average, took longer than the \textit{baseline} condition to get the participant to accept an offer with a large effect size ($G=0.77$) shown using the Hedges' G~test. The \textit{Impatient Low-arousal} condition however, needed fewer interactions than the baseline condition with a medium effect size ($G=0.44$) in the other direction. Comparing the two measured conditions directly thus, shows a large effect size ($G=0.87$) for the number of interactions. Furthermore, under the \textit{Impatient Low-arousal} condition, the robot was able to reach an offer $>=50\%$ of the points for $80\%$ of the participants as compared to $62\%$ for the \textit{Patient High-arousal} condition. Despite reaching a higher offer more often, the success rate and the mean accepted offer for the \textit{Impatient Low-arousal} condition were lower than those for the \textit{Patient High-arousal} condition. As participants increasingly received more points in the \textit{Impatient Low-arousal} condition, they frequently exhausted the $20$ offers, anticipating the robot to increase the offer further. This observation is validated by the \textit{mean offer value} before aborting being higher for the \textit{Impatient Low-arousal} condition with a large effect size ($G>2.0$) between the two measured conditions.

\subsubsection{Qualitative Results}
\label{sec:quality}
Since the participants' subjective evaluation of the robot's negotiation strategy influences their acceptance or rejection, quantitative factors provide only partial information about the robot's overall performance. Thus, participants' evaluations on the $3$-part Likert-scale questionnaire, based on the GODSPEED~\cite{Bartneck2008Measurement}, Mind Perception~\cite{Gray2007} and Asch's Personality Impression tests~\cite{asch1946forming}, are examined. 

As participants evaluate each measured condition with respect to the \textit{baseline} (No Core) condition, the two measured conditions can be compared directly only if the baseline sub-condition is evaluated the same way in the two groups. A Two-sided Mann-Whitney~U test shows no significant difference ($p>0.05$) in any dimension between the two baselines in any questionnaire. This allows for the two \textit{measured} sub-conditions to be compared to each other, directly. 

\begin{enumerate}[label=(\alph*),leftmargin=0.5cm]
    \item \textbf{GODSPEED:} The GODSPEED test~\cite{Bartneck2008Measurement} is used to measure participants' impression of the robot on anthropomorphism, animacy, likeability, perceived intelligence and perceived safety. A one-sided Mann-Whitney~U test is conducted for all dimensions with an alternative hypothesis that the \textit{Impatient Low-arousal} condition is rated higher than \textit{Patient High-arousal}. The results show no significant differences $(p>0.05)$ in any dimension despite some evidence for the robot rated as more \textit{natural} $(U=154.5$, $p=0.07)$, \textit{human-like} ($U=158.0$, $p=0.053)$ and \textit{conscious} $(U=158.0$, $p=0.061)$ under the \textit{Impatient Low-arousal} condition.

    \item \textbf{Mind Perception:} The Mind Perception test~\cite{Gray2007} measures \textit{agency} and \textit{experience} for attributing a \textit{mind} in an entity (in this case, \ac{NICO}). The robot is evaluated on its ability to experience \textit{fear}, exercise \textit{self-control}, feel \textit{pleasure}, \textit{remember} the participant, feel \textit{hunger} and to \textit{act morally}. Based on these factors, the robot's \textit{agency} and \textit{experience} under different conditions is concluded. A one-sided Mann-Whitney~U test is conducted with the alternative hypothesis that the \textit{Impatient Low-arousal} condition is rated higher on \textit{agency} and \textit{experience} with no significant difference $(p>0.05)$ concluded between the two conditions.

\begin{figure}
\centering
\vspace{-3.5mm}\includegraphics[width=0.5\textwidth]{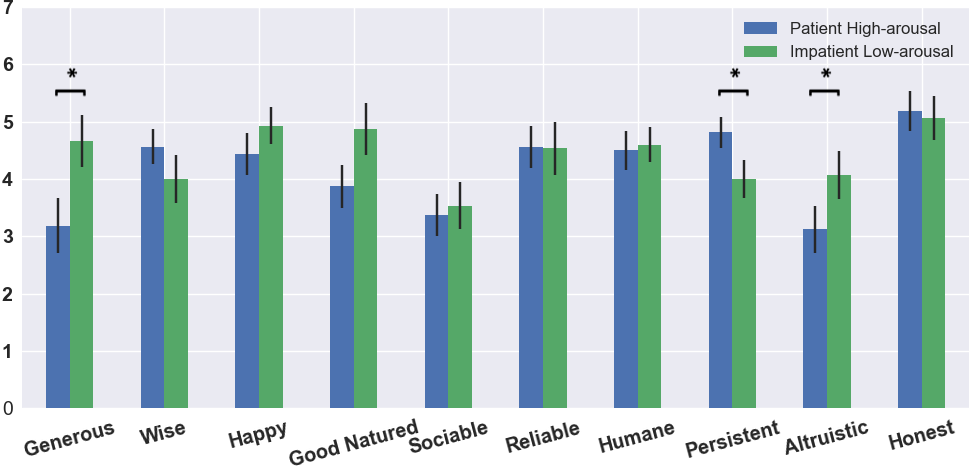}

\caption{Asch's Test results with mean and $95\%$ CI for individual dimensions comparing the two measured conditions.}
\label{fig:asch}
\vspace{-3mm}
\end{figure}

    \item \textbf{Asch's Formation of Impressions of Personality:} Asch's study~\cite{asch1946forming} measures the impact of independent behavioural traits on the overall impression of an individual. Here, participants evaluate \ac{NICO} on $10$ different parameters. Their impressions for the robot under the two measured conditions can be seen in Fig.~\ref{fig:asch}. For all dimensions, except \textit{wisdom} and \textit{persistence}, the \textit{Impatient Low-arousal} condition is rated higher, while in these dimensions, the \textit{Patient High-arousal} condition is rated higher. A one-sided Mann-Whitney~U test conducted on all dimensions shows significant results $(p<0.05)$ in the \textit{generous} $(U=67.0$, $p=0.018)$ and \textit{altruistic} $(U=74.0$, $p=0.033)$ dimensions in favour of the \textit{Impatient Low-arousal} condition (see Table.~\ref{tab:asch}), and in the \textit{persistence} $(U=75.0$, $p=0.034)$ dimension in favour of the \textit{Patient High-arousal} condition. Despite some evidence supporting the alternative hypothesis for \textit{good-natured} dimension $(U=79.0$, $p=0.052)$, no other conclusions can be drawn.
\end{enumerate}

\vspace*{-2mm}
\section{Discussion}
\label{sec:discussion}

This work explores a robot's appraisal to ground evolving affective representations that not only consider the behaviour of the participant during an interaction (see Section~\ref{sec:perception}), but also understand its impact on the conversation (see Section~\ref{sec:mood}), learning how to respond to them (see Section~\ref{sec:behavemodel}). This is guided by the personality traits of the robot (see Section~\ref{sec:core}) which have a significant impact on its affective appraisal. 

Quantifying the affective impact of the duration of an interaction is beneficial for a robot, particularly in collaborative \ac{HRI} scenarios. A \textit{patient} time perception can be helpful in dealing with \textit{negative} or even \textit{aggressive} situations as it will allow the robot to maintain a positive outlook during the interaction. This can be beneficial for robots acting as companions for humans in different collaborative scenarios such being caretakers for the elderly and teachers for the young. Conversely, \textit{impatience} results in a significantly lower intrinsic state of the robot, rapidly decaying its mood as the interaction progresses. This may enhance spontaneity in robot behaviour as it finds ways to resolve a negotiation quickly, to avoid negative intrinsic states. 


Interactions with high intensity cause the robot to form \textit{excitatory} (or high-arousal) tendencies that amplify its affective state. While interacting with the users, the robot is easily excitable, experiencing every situation in the extreme. An \textit{inhibitory} (or low-arousal) conditioning, on the other hand, results in a subjugated behaviour of the robot, diminishing the impact of affective interactions and adopting an inert approach towards its interaction with the users.

\begin{table}
\centering
\footnotesize{
\caption{One-sided Mann-Whitney~U test for Asch's Test with alternative hypothesis that the Impatient Low-arousal condition is rated higher.}
\vspace{-1mm}

\label{tab:asch}
\setlength\tabcolsep{3.5pt}
\begin{tabular}{lrr|lrr}
\toprule
\makecell[l]{\textbf{Dimension}}  &   \makecell[c]{\textbf{U-Statistic}}    &   \makecell[c]{\textbf{p-value }}   & \makecell[l]{\textbf{Dimension}}  &   \makecell[c]{\textbf{U-Statistic}}    &   \makecell[c]{\textbf{p-value }}\\
\midrule
\textbf{Generous}   &   \textbf{67.0}   &   \textbf{0.018}  &   Reliable            &   120.5           &   0.516 \\  
Wise                &   142.5           &   0.825           &   Humane              &   119.0           &   0.492 \\
Happy               &   95.0            &   0.161           &   \emph{\textbf{Persistent}} &   \textit{\textbf{165.0}}  &   \textit{\textbf{0.969}} \\
\textit{Good Natured }       &   \textit{79.0 }           &   \textit{0.052  }         &   \textbf{Altruistic} &   \textbf{ 74.0}  &   \textbf{ 0.033} \\ 
Sociable            &   112.5           &   0.387           &   Honest              &   123.5           &   0.564 \\
\bottomrule
\end{tabular}
\vspace{-4mm}
}

\end{table}
Combining time perception and social conditioning allows for modelling specific personality dispositions in the robot with the two influences either complementing each other, for example, \textit{Patient high-arousal} and \textit{Impatient low-arousal}, or contrasting each other, for example, \textit{Patient low-arousal} and \textit{Impatient high-arousal} conditions. These conditions have a distinct impact on the affective appraisal of the robot (see Table~\ref{tab:result_affcore_bias}) as the resultant mood does not merely mimic the user's affective state but reflects the robot's intrinsic dispositions.

The robot's intrinsic mood, modulated by specific \textit{affective core} dispositions, as well as history with a user, governs its negotiations in the Ultimatum Game. In our experiments, the \textit{patient high-arousal} robot is witnessed to stand its ground longer, driving a hard bargain with users while the \textit{impatient low-arousal} robot, on the other hand, is more giving and generously offers more points. This is highlighted in the quantitative analysis (see Section~\ref{sec:behaveresults_quan}) of the robot as well as the subjective evaluations by the participants (see Section~\ref{sec:quality}). 

During interactions, based on the robot's behaviour, the participants were witnessed adopting different negotiating strategies. While some approached the interaction donning a more \textit{commanding} role, strongly arguing with the robot to yield, others followed a fawning approach trying to manipulate the robot by smiling more often and \textit{requesting} more points. Both strategies, given the experiment condition and the expressiveness of the participants, worked to some extent with the robot offering as high as $52\%$ of the points. Furthermore, at the beginning of the interactions, some participants were more \textit{conscious} and \textit{distant}, but as the interaction progressed, they became more \textit{open} and \textit{proactive} in the interaction. This is seen in the reasoning provided by them for their rejection which ranged from a cold and direct ``\textit{I want more points}'' later to a more expressive and layered \textit{``Come on, NICO. This isn't fair. You can do better''}. This suggests that as the interaction progressed, the robot was able to \textit{engage} the users. It exhibited responsiveness towards the users' negotiating strategies, initially yielding to their demands for more points but, as the interaction progressed, it adapted its negotiation strategy, encouraging the users to also adapt.

Despite the participants noticing significant differences in its negotiating strategy (see Table~\ref{tab:asch}), the general perception of the robot did not change under different conditions. This could be due to the fact that the only difference between conditions is in how the robot updates its offers. The interaction structure, what is said and robot's facial expressions remain the same between conditions. This difference is perhaps \textit{too subtle} to induce an overall change in perception towards the robot. It will be interesting to also modulate dialogues to reflect the robot's mood, adding phrases that reflect the \textit{affective core} condition. 

\section{Conclusion}
\label{sec:conclusion}

In this work, we present a comprehensive framework for modelling personality-driven robot behaviour in collaborative \ac{HRI} scenarios. Using a multi-modal affective appraisal model, it forms an evolving understanding of human behaviour, yielding intrinsic responses in the robot towards the user, that constitute its own affective state. This intrinsic state is used to learn negotiating behaviour in the Ultimatum Game. The \textit{affective core} of the robot realises specific personality traits in the robot that influence its intrinsic state as well as its behaviour. This is beneficial for the robot to dynamically interact with users rather than following static pre-determined behaviour policies. 

The results from the user study show that the participants were able to notice the effect of the \textit{affective core} on factors such as \textit{generosity} and \textit{persistence} which directly evaluated the robot's behaviour in the Ultimatum Game. The general impression of the robot, however, did not change significantly. Further experimentation is needed, involving longitudinal studies with more participants, to conclude any significant impact on the overall impression of the robot. Furthermore, in the user study, the \textit{affective core} models are pre-trained and used after freezing the weights for the Gamma-\ac{GWR} model. This was done to simplify the training and eliminate the effect of changing \textit{affective core} biases on the performance of the robot. It will be interesting to let these models to grow and adapt as the robot interacts with more users, allowing the robot to change its outlook on the users as it interacts with them.

\begin{acronym}
\acro{Aff-Wild}{Affect-in-the-Wild}
\acro{AFEW-VA}{Acted Facial Emotions in-the-wild for Valence and Arousal}
\acro{AU}{Action Units}
\acro{BML}{Behavioural Markup Language}
\acro{BMU}{Best Matching Unit}
\acro{CHL}{Competitive Hebbian Learning}
\acro{CCC}{Concordance Correlation Coefficient}
\acro{CNN}{Convolutional Neural Networks}
\acro{DDPG}{Deep Deterministic Policy Gradients}
\acro{DPG}{Deterministic Policy Gradients}
\acro{DQN}{Deep Q Networks}
\acro{EIQ}{Emotional Intelligence}
\acro{eNTERFACE'05}{eNTERFACE'05 Audio-Visual Emotion}
\acro{FACS}{Facial Action Coding System}
\acro{FER}{Facial Expression Recognition}
\acro{FML}{Functional Mark-up Language}
\acro{GNG}{Growing Neural Gas}
\acro{GWR}{Growing-When-Required}
\acro{HHI}{Human-Human Interaction}
\acro{HRI}{Human-Robot Interaction}
\acro{KT}{Knowledge Technology}
\acro{MCCNN}{Multi-Channel Convolutional Neural Network}
\acro{MDP}{Markov Decision Process}
\acro{MFCC}{Mel-Frequency Cepstral Coefficients}
\acro{MLP}{Multilayer Perceptron}
\acro{NICO}{Neuro-Inspired Companion}
\acro{OMG-Emotion}{One Minute Gradual-Emotion}
\acro{PAD}{Pleasure-Arousal-Dominance}
\acro{RAVDESS}{Ryerson Audio-Visual Database of Emotional Speech and Song}
\acro{ReLU}{Rectified Linear Unit}
\acro{RL}{Reinforcement Learning}
\acro{SAIBA}{Situation-Agent-Intention-Behaviour-Action}
\acro{SAVEE}{Surrey Audio-Visual Expressed Emotion}
\acro{SER}{Speech Emotion Recognition}
\acro{(DE)SIRE}{Description of Emotion through Speed, Intensity, Regularity and Extent}
\acro{SOM}{Self-Organising Map}
\acro{TD}{Temporal Difference}
\acro{TORCS}{The Open Racing Car Simulator}
\end{acronym}
\balance

\bibliographystyle{IEEEtran}
\bibliography{main.bib}
\vspace{-0.12in}
\begin{IEEEbiography}[{\vspace*{-6mm}\includegraphics[width=1in,height=1in,clip,keepaspectratio]{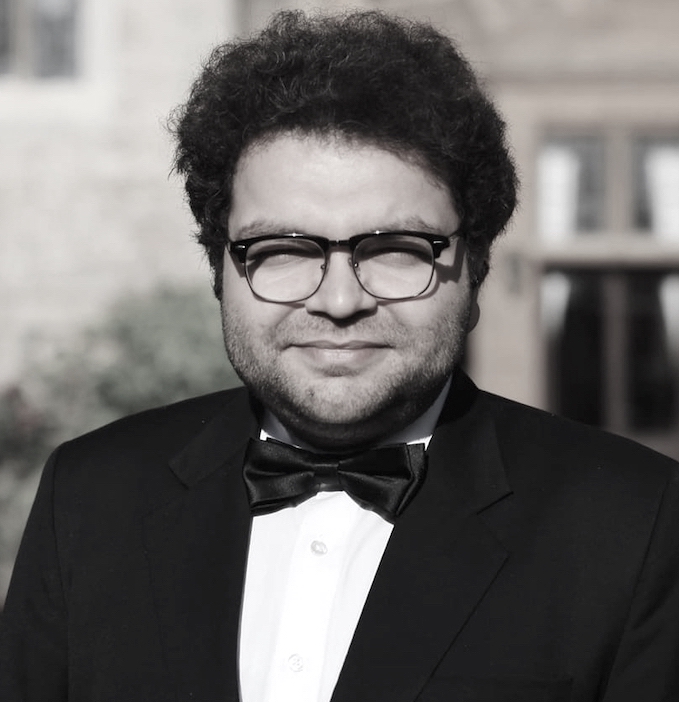}}]{Nikhil Churamani} received his MSc. degree with distinction in Intelligent Adaptive Systems from Universit\"at Hamburg in 2018. He is currently a doctoral student at the Department of Computer Science and Technology, University of Cambridge, UK. His research interests include Neural Networks, Computer Vision, Continual Learning and Human-Robot Interaction. His current research investigates deep learning models for lifelong learning in social robots focused on Human-Robot Interaction and affect-driven behavioural learning.
\end{IEEEbiography}
\vspace{-0.3in}
\begin{IEEEbiography}[{\vspace*{-6mm}\includegraphics[width=1in,height=1in,clip, keepaspectratio]{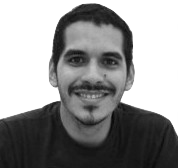}}]{Pablo Barros} received his BSc. in Information Systems from the Universidade Federal Rural de Pernambuco and his MSc. in Computer Engineering from the Universidade de Pernambuco, both in Brazil. He received his Ph.D. in Computer Science from Universit\"at Hamburg, Germany. Currently, he is a research scientist at the Italian Institute of Technology. His main research interests include neural networks and their applications to affective and social robots. He has been a guest editor of the journals IEEE Transactions on Affective Computing, Frontiers on Neurorobotics and Elsevier Cognitive Systems Research. He is also the publication chair of ICDL-EpiRob 2020.
\end{IEEEbiography}
\vspace{-0.2in}
\begin{IEEEbiography}[{\vspace*{-8mm}\includegraphics[width=1in,height=1in,clip,keepaspectratio]{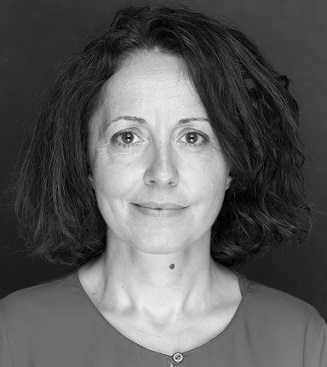}}]{Hatice Gunes}(Senior Member, IEEE) received the Ph.D. degree in computer science from the University of Technology Sydney, NSW, Australia. She is a Reader with the Department of Computer Science and Technology, University of Cambridge, U.K., leading the Affective Intelligence and Robotics Lab. Her expertise is in the areas of affective computing and social signal processing cross-fertilizing research in human behaviour understanding, computer vision, signal processing, machine learning, and human–robot interaction. She has published over 100 papers in the above areas. Dr Gunes is the former President (2017-2019) of the Association for the Advancement of Affective Computing, was the General Co-Chair of ACII 2019, and Program Co-Chair of ACM/IEEE HRI 2020 and IEEE FG 2017. Her research has been supported by various competitive grants, with funding from EPSRC, Innovate U.K., British Council, and EU Horizon 2020. She is a Fellow of the Engineering and Physical Sciences Research Council, U.K., a Faculty Fellow of the Alan Turing Institute, and a Staff Fellow of Trinity Hall Cambridge.
\end{IEEEbiography}
\vspace{-0.2in}
\begin{IEEEbiography}[{\vspace*{-6mm}\includegraphics[width=1in,height=1in,clip,keepaspectratio]{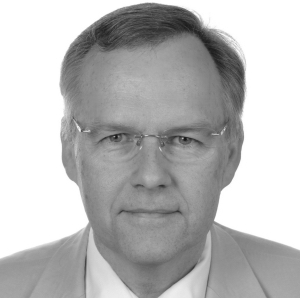}}]{Stefan Wermter} is Full Professor at the University of Hamburg, Germany, and Director of the Knowledge Technology Institute at the Dept. of Informatics. His main research interests are in the fields of neural networks, hybrid knowledge technology, neuroscience-inspired computing, cognitive robotics, and human-robot interaction. He has been associate editor of Transactions on Neural Networks and Learning Systems, and is associate editor of Connection Science, and International Journal for Hybrid Intelligent Systems. He is on the editorial board of journals Cognitive Systems Research, Cognitive Computation and Journal of Computational Intelligence. Currently, he is co-coordinator of the international collaborative research centre on Crossmodal Learning~(TRR-169) and coordinator of the European Training Network SECURE on safety for cognitive robots. In 2019 he has been elected as the President for the European~Neural~Network~Society~2020-2022.

\end{IEEEbiography}

\end{document}